\documentclass[final,5p,times,twocolumn]{elsarticle}

\usepackage{amsmath}
 \usepackage{multirow}
 \usepackage{graphics}
 \usepackage{subcaption}
 \usepackage{bm}
 \usepackage{booktabs}
 \usepackage{makecell}
 \usepackage{algorithm}
\usepackage{algpseudocode}
\usepackage{comment}
\usepackage{amssymb}

\usepackage{enumitem}

\usepackage{colortbl}

\usepackage{setspace}

\let\Algorithm\algorithm
\renewcommand\algorithm[1][]{\Algorithm[#1]}

\begin{document}

\begin{frontmatter}

\title{Hierarchical Multi-resolution Mesh Networks for Brain Decoding}

\author[mymainaddress]{Itir Onal Ertugrul} 
	\cortext[cor1]{Corresponding author:} 
	\ead{itir@ceng.metu.edu.tr}
	
    \author[mysecondaryaddress]{Mete Ozay}
    
	\author[mymainaddress]{Fatos Tunay Yarman-Vural}
	
    \address[mymainaddress]{Department of Computer Engineering, Middle East Technical University, Ankara, Turkey}
    
	\address[mysecondaryaddress]{Graduate School of Information Sciences, Tohoku University, Sendai, Miyagi, Japan}





\begin{abstract}
We propose a new  framework, called Hierarchical Multi-resolution Mesh Networks (HMMNs), which establishes a set of brain networks at multiple time resolutions of fMRI signal to represent the underlying cognitive process. The suggested framework, first, decomposes the fMRI signal into various frequency subbands using wavelet transforms. Then, a brain network, called mesh network, is formed at each subband by ensembling a set of local meshes. The locality around each anatomic region is defined with respect to a neighborhood system based on functional connectivity. The arc weights of a mesh are estimated by ridge regression formed among the average region time series. In the final step, the adjacency matrices of mesh networks obtained at different subbands are ensembled for brain decoding under a hierarchical learning architecture, called, fuzzy stacked generalization (FSG). Our results on Human Connectome Project task-fMRI dataset reflect that the suggested HMMN model can successfully discriminate tasks by extracting complementary information obtained from mesh arc weights of multiple subbands. We study the topological properties of the mesh networks at different resolutions using the network measures, namely,  node degree, node strength, betweenness centrality and global efficiency; and investigate the connectivity of anatomic regions, during a cognitive task. We observe significant variations among the network topologies obtained for different subbands. We, also, analyze the diversity properties of classifier ensemble, trained by the mesh networks in multiple subbands and observe that the classifiers in the ensemble collaborate with each other to fuse the complementary information freed at each subband. We conclude that the fMRI data, recorded during a cognitive task, embed diverse information across the anatomic regions at each resolution. 

\end{abstract}


\end{frontmatter}


\section{Introduction}

Traditional brain decoding methods employ the activation of voxels  for pattern analysis of functional magnetic resonance imaging (fMRI) data \cite{Kamitani2005,Cox2003}. However, recent studies show that brain networks, which are formed by correlating fMRI signals obtained from voxel pairs, provide more information compared to the temporal dynamics of voxels for brain decoding \cite{Lindquist2008}. 

There has been a shift in brain decoding paradigms towards modeling the brain connectivity by networks, since they offer a proper framework to recognize brain patterns and represent the interactions among regions \cite{Richiardi2013}. Richiardi et al.~\cite{Richiardi2011}, \cite{Richiardi2013} suggest various descriptors  to extract features from fMRI connectivity graphs and decode cognitive states using the descriptors extracted from these graphs .  Fornito et al. \cite{Fornito2012} propose a method to learn brain connectivity models using edge connections computed between graph nodes. Shirer et al. \cite{Shirer2011} employ functional connectivity models to decode continuous and free streaming cognitive states. Ekman et al. \cite{Ekman2012} analyze adjustments in functional connectivity models from a graph theoretical perspective. These studies propose methods to recognize connectivity patterns by modeling various types of pairwise relationships of nodes of connectivity graphs.

Unlike the methods which estimate pairwise connectivity between the voxels, Onal et al. \cite{Onal2015_1} propose a method which forms connectivity graphs by ensembling a set of local meshes. In this graph representation, the nodes correspond to the voxels. A node is connected to its p-nearest neighboring voxels to form a star mesh, where a neighborhood of a voxel is defined with respect to Euclidean distance between voxel coordinates (called spatial neighborhood), or functional similarities between voxel BOLD responses (called functional neighborhood). They represent the BOLD response recorded at each voxel (node) as a linear combination of the BOLD responses of its neighboring voxels. The arc weights of each mesh are then estimated by Ridge regression to represent the relationship among the voxels within their spatial \cite{Onal2015_1} or functional \cite{Onal2015_2} neighborhood. Finally, they embed the arc weights of local meshes into a feature vector to train a classifier for brain decoding. This approach, which aggregates the locally connected meshes under a global network model, resulted in better decoding performances, compared to pairwise relationship models. 

It is known that brain processes information in multiple frequency bands, and different frequencies of neuronal activity have been linked to the BOLD signal \cite{Thompson2015}. Features of spoken sentences, visual stimuli or development of social interaction may unfold over distinct time scales \cite{Kauppi2014}. Kauppi et al. \cite{Kauppi2010} report that distinct regions exhibit inter-subject correlations at low, medium or high frequencies. These studies imply that different regions of the brain  discriminate the conditions in different subbands. Therefore, multi-resolution analysis of fMRI signal is crucial for analyzing and decoding the cognitive states.

Wavelet transforms are widely used to represent the fMRI signals in multiple resolutions with approximately decorrelated coefficients \cite{Vandeville2006}.  Bullmore et al. \cite{Bullmore2004} reported that the brain has fractal property (also called $1/f$-like property),  where the statistical properties that describe the structure of a system, in time or space, do not change over a range of different scales \cite{Mandelbrot1977}. In this sense, wavelets are well-suited for multi-resolution fMRI analysis \cite{Xu2002}, \cite{Dinov2005}. Adaptivity of wavelets to local or non-stationary features of an fMRI signal makes them suitable choices for the analysis of the fMRI signal, which is expected to include non-stationary features of interest at several scales \cite{Bullmore2004}. Furthermore, Discrete Wavelet Transform (DWT) has decorrelating capability for a wide class of signals having $1/f$-like property. In other words, even if the data is highly correlated, the correlations computed between wavelet coefficients are generally small.

In this study, we assume that fMRI signals, which are reconstructed by the decorrelated wavelet coefficients for different time resolutions, carry complementary information in the corresponding feature spaces. This assumption is supported by the study of Richiardi et al. \cite{Richiardi2011} which show that the multi-resolution signals obtained with an orthogonal DWT are quasi class-conditionally independent. Therefore, we expect that the classifiers trained using multi-resolution signals are diverse, and fusion of their decisions would yield high performance for decoding the brain signals. This approach requires a fast and efficient decision fusion method to ensemble the classifiers trained by the fMRI signals at different resolutions.

Ensemble learning classifiers are used in many fMRI studies, including brain decoding. In a pioneering study by Kuncheva et al. \cite{Kuncheva2010}, multiple classifiers are trained by the various subsets of  samples. Recently, Alkan et al. \cite{Alkan2015} partition the brain into homogeneous regions with respect to functional similarity of voxel time series and train a different classifier at each region. Multiple classifiers are also trained by a set of complementing stimuli \cite{Cabral2012} or subbands \cite{Richiardi2011}. Then, the results of the classifiers are ensembled by majority voting techniques. These ensemble learning methods have been shown to outperform the methods based on a single classifier. 
A hierarchical ensemble learning method, suggested by  Ozay et al.\cite{Ozay2016}, fuses the decisions of multiple classifiers  by a meta classifier. This method, called, fuzzy stacked generalization (FSG), is shown to outperform  a number of ensemble learning methods \cite{Ozay2016} in Multi Voxel Pattern Analysis (MVPA). 


\begin{figure*}[t]
	\centering
	\includegraphics[scale=0.5]{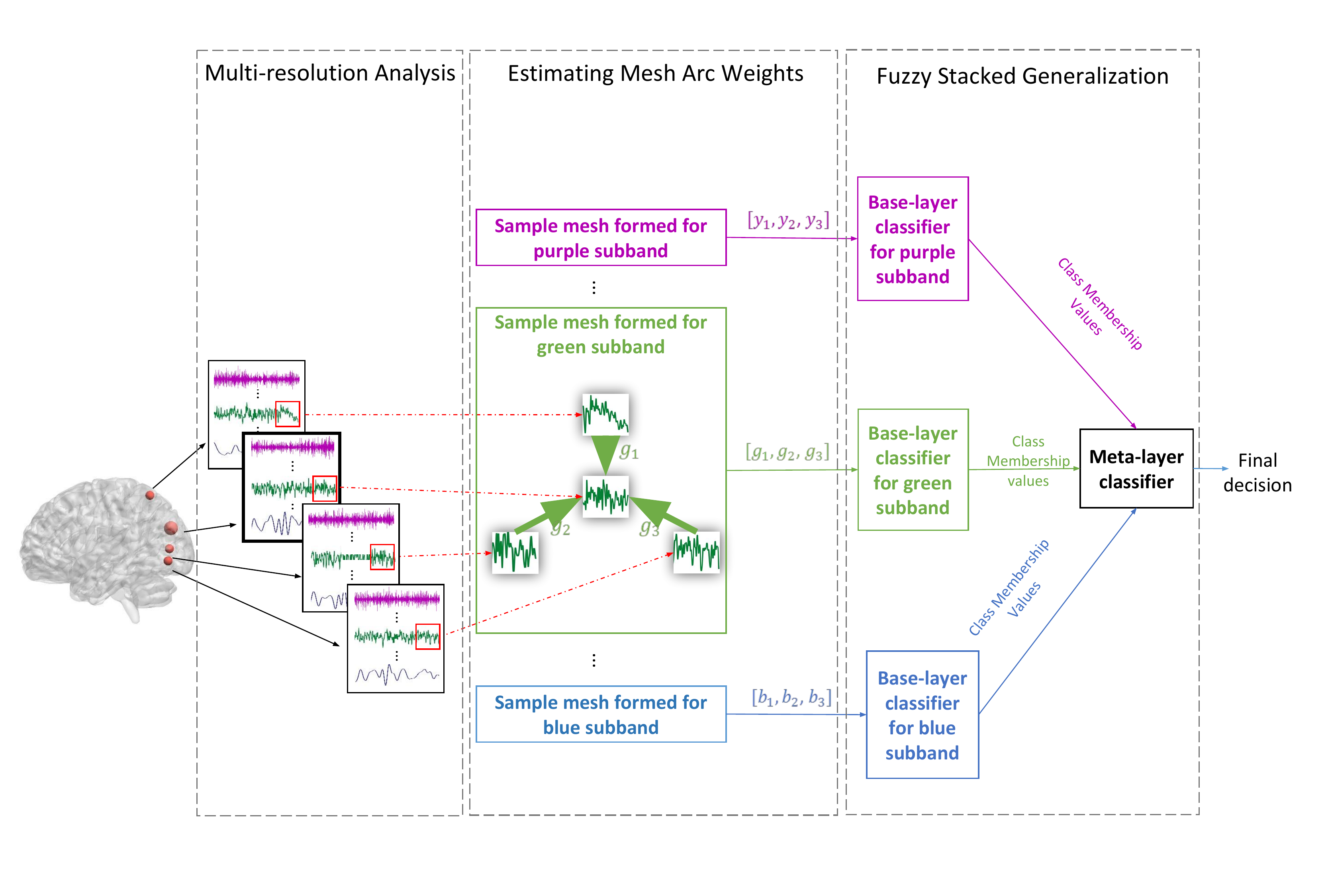}
	\caption{
	Overview of the proposed hierarchical framework (HMMNs). A random seed region and its three functionally nearest neighbors (with smaller sizes) are denoted on the template brain. In the multi-resolution analysis step, the average fMRI signal of each region is decomposed into a a set of coarse-to-fine signals. Then, the meshes are formed around each seed region for all subbands, and mesh arc weights are estimated. Finally, the estimated arc weights of the meshes are ensembled under a network structure and fed as an input to the hierarchical ensemble learning architecture, called, FSG, where the final decision is computed by a meta-layer classifier. }
	\label{fig:flowchart}
\end{figure*}

In this study, we propose a framework, called, Hierarchical Multi-resolution Mesh Networks (HMMNs), which fuses a set of multi-resolution brain networks by a hierarchical learning architecture and enables us to analyze the network topology of brain in multiple resolutions (see Figure \ref{fig:flowchart}). The suggested framework consists of the following steps:
\begin{enumerate}[leftmargin=0cm,itemindent=.5cm,labelwidth=\itemindent,labelsep=0cm,align=left]

\item We estimate a representative time series for each anatomic region by simply taking the average of all the voxel time series in that region. While this approach reduces the effect of the noise in voxel intensity values, it also reduces the dimension of voxel space to the dimension of the region space.

\item  We decompose the representative time series of an anatomic region into a set of multi-resolution signals using wavelet transform. The multi-resolution representation of an anatomic region enables us to observe the essential information and the underlying cognitive task processed in this region.

\item For each time resolution, we estimate an independent network, where the nodes correspond to the anatomic regions and the arc weights represent the local relationship of an anatomic region with its neighbors.  The brain network is defined as an ensemble of meshes formed in the functional neighborhood of the anatomic regions which has been shown to perform slightly better than spatial neighborhood in \cite{Onal2015_2}. The arc weights of each mesh are estimated using ridge regression.

\item We adopt the two-layer FSG architecture for fusion of the brain networks obtained in multiple resolutions. At the base-layer of FSG, a set of  logistic regression classifiers is trained by the brain networks at different resolutions. Then, the class posterior probabilities, obtained at the output of logistic regression classifiers are concatenated under a decision space to form the input of a meta-layer classifier, which is trained for final decision.
\end{enumerate}
We perform subject-transfer learning in which the classifiers are trained by the data obtained from a group of subjects and tested by the data obtained from another group. Our results reflect that the arc weights of mesh networks, which represent the connectivity among the regions, provide better graph embeddings to classify the cognitive tasks compared to pairwise correlations between regions or raw fMRI data. Furthermore, training  classifiers by mesh arc weights obtained from different time resolutions, and fusing their decisions leads to better classification accuracy compared to training a single classifier by the mesh weights of original and single-resolution signal.

We also analyze the node degree, node strength, betweenness centrality and global efficiency measures of the mesh networks obtained for each subband, and investigate their relationship to the classification performance. We observe significant diversity among the networks obtained for different subbands for each measure, which shows that the signals  embed different characteristics of brain connectivity at each resolution. 

Finally, we investigate the class discrimination power of the mesh networks in each resolution. Our brain decoding results reflect that the classifiers, trained by the multi-resolution networks  collaborate with each other to complement the information embedded at different subbands. HMMN model provides higher decoding performances compared to single resolution methods and multi-resolution methods based on pairwise correlation networks.


\section{Data Acquisition and Experimental Setup}

We use 900 subject release of task fMRI data from Human Connectome Project (HCP).  During the experiments, the subjects performed seven tasks namely Emotion Processing \cite{Hariri2002}, Gambling \cite{Delgado2000}, Language \cite{Binder2011}, Motor \cite{Buckner2011}, Relational Processing \cite{Smith2007}, Social Cognition \cite{Castelli2000,Wheatley2007}, Working Memory (WM). Further information about experiments can be found in \cite{Barch2013}. We use data obtained from $S = 808$ subjects who performed all of the seven cognitive tasks. A task experiment session, $q$, yields the fMRI recordings with total of $D_q$ scans (brain volumes),  to represent the underlying task. The duration and the number of scans, $D_q$, vary for each task experiment. However, the duration of the experiment for a particular task is the same for all participants (see Table \ref{tab:7tasks}). 

\begin{table}[ht]
	\centering
	\caption{Number of scans per task experiment session and the duration for each task (min:sec).}
	\label{tab:7tasks}
    \resizebox{\columnwidth}{!}{%
	\begin{tabular}{|l|c|c|c|c|c|c|c|}
		\hline
		& \textbf{Emotion} & \textbf{Gambling} & \textbf{Language} & \textbf{Motor} & \textbf{Relational} & \textbf{Social} & \textbf{WM} \\ \hline
		\textbf{Scans}    & 176              & 253               & 316               & 284            & 232                 & 274             & 405         \\ \hline
		\textbf{Duration} & 2:16             & 3:12              & 3:57              & 3:34           & 2:56                & 3:27            & 5:01        \\ \hline
	\end{tabular}}
\end{table}

We use $R = 90$ anatomical regions of 116 AAL, after removing the anatomical regions in Cerebellum and Vermis. The voxel coordinates and size of each region are the same for all subjects. The fMRI signal recorded during a task experiment session, $q$, consists of a set of voxel time series of length $D_q$ to represent the underlying cognitive task. The time series, called BOLD signal, obtained at each voxel $v$ is represented by a function of time $t$ with $X_v(t)$. Then, we obtain a representative time series $X_r(t)$ for each region $r$ by spatially averaging the time series of voxels residing in that region using
\begin{equation}
\label{eq:average_sig}
X_r(t) = \dfrac{1}{V_r} \sum_{\forall v \in r} X_v(t) ,
\end{equation}
where $V_r$ denotes the number of voxels residing in region $r$. Note that the length $D_q$ of $X_r(t)$, depends on the number of time scans of a task experiment session $q$.

\section{{Hierarchical Multi-Resolution Mesh Networks (HMMNs)}}

The major goal of the suggested HMMNs model is to represent a cognitive task by a set of course-to-fine brain networks, each of which is expected to signify complementing information about the underlying brain activities. This model enables us to study the activities and interactions of brain regions in  multiple resolutions during a cognitive states. HMMNs also enable us to boost the brain decoding performance due to the fusion of diverse connectivity structures represented in mesh networks.

In the following subsections, first we explain the multi-resolution representation of fMRI data. Next, we represent each cognitive task by a set of multi-resolution brain networks.  Finally, we explain a two-layer learning architecture, called fuzzy stacked generalization (FSG), which fuses the brain networks, obtained in different resolution for brain decoding.

\subsection{Multi-resolution Representation of fMRI Signals}
\label{sec:multiresolutional}

The first step for construction of the proposed HMMNs is to decompose the fMRI recordings into a set of signals in different time resolutions. The multi-resolution representation of fMRI data enables us  to estimate and analyze the time scale(s) in which the anatomical regions process a particular information (\cite{Richiardi2011, Thompson2015,Kauppi2014,Kauppi2010}). Discrete Wavelet Transform (DWT) is preferred for multi-resolution analysis of fMRI data due to its decorrelating property \cite{Fan2003} and adaptivity to non-stationary features of fMRI signals \cite{Bullmore2004}.

The average signal $X_r(t)$, in \eqref{eq:average_sig}, represents the information processed in anatomical region, $r$. As the next step, we decompose all the representative signals, $X_r(t)$, for $r = 1, 2, \ldots ,R$, into a set of signals using subband coding along the temporal dimension to analyze the regional brain activities at different resolutions. In multi-resolution analysis step of HMMNs, shown in Figure \ref{fig:flowchart}, we define the mother wavelet by, 
\begin{equation}
\Psi_{l,k}(t) = \frac{1}{\sqrt{2^l}} \Psi \left( \frac{t-2^l k}{2^l} \right), \forall t=1,2,\ldots,T,
\end{equation}
where $\int \psi(t)dt = 0$, $\forall l=1,2,\ldots L$ and $\forall k=1,2,\ldots, K$. Similarly, we define the father wavelet that satisfies ${\int \phi(t)dt = 1}$ by
\begin{equation}
\Phi_{l,k}(t) = \frac{1}{\sqrt{2^l}} \Phi \left( \frac{t-2^l k}{2^l} \right), \forall t, l, k.
\end{equation}
There are total of $L$ decomposition levels where ${l = 1, 2, \ldots, L}$. Also, $k$ represents the location to which the wavelet is translated in time, where $k = 1, 2, \ldots, K$. The value of $K$ changes for each $l$ such that $K = T/2^l$, and $T$ denotes the duration of the signal. Note that each high or low frequency subband is spanned by translated versions of a single mother wavelet or father wavelet, respectively.

We apply discrete wavelet transform (DWT) to the average fMRI signal $X_r(t)$ obtained from a region $r$ to decompose the signal into two sets of orthogonal components, namely, a set of approximation coefficients $\mathcal{A} = \{ \alpha_{r,l,k}\}$, and a set of detail coefficients $\mathcal{D}=\{ \delta_{r,l,k}\}$. While the detail coefficients represent the information about variation in the data at a given scale, the approximation coefficients contain the residual of the signal after the information in the provided scale and all finer scales are removed \cite{Bullmore2004}. Then, we can recover the original signal $X_r(t)$ obtained from region $r$ by
\begin{equation}
\label{eq:recover_signal}
X_r(t) = \sum_k \alpha_{r,l,k} \Phi_{l,k}(t) + \sum_{l' \leq l} \sum_k \delta_{r,l',k}\Psi_{l',k}(t) .\end{equation}

As it can be seen from \eqref{eq:recover_signal}, we can recover the original signal by first multiplying the father wavelet with the approximation coefficient at a particular decomposition level $l$, and then adding them to the multiplication of mother wavelet and detail coefficients of all levels greater than or equal to this particular level, $\forall l' \leq l$. 

We can also reconstruct only the approximation or detail parts of the signal at a decomposition level. For example, if we decompose the signal in the first level, then we can reconstruct the approximation, $A_1$, and detail, $D_1$, parts using inverse discrete wavelet transform. If we continue decomposing the approximation part $A_1$ one more level, then, we obtain the second level approximation $A_2$ and detail $D_2$ parts. Therefore, for a given signal, we can obtain various signal components, each of which contains the information about components of the signal in different time resolutions.

Let $j \in \mathcal{J} = \{0, 1, \ldots, 2L\}$ be the index of the entries of the ordered set of subbands $\{A_0, A_1, \ldots , A_L, D_1, \ldots, D_L\}$. Then, we can reconstruct a particular approximation or detail part of a signal using approximation and detail coefficients, respectively, as follows:
\begin{equation}
\label{eq:recover_parts}
x_{j,r}(t)= \begin{cases}
X_r(t), &\text{if } j = 0\\
\sum_k \alpha_{r,l,k} \Phi_{l,k}(t) \text{ where } l = j,   & \text{if } 1 \leq j\leq L\\
\sum_k \delta_{r,l,k}\Psi_{l,k}(t)  \text{ where } l = j-L+1,  & \text{if } j > L
\end{cases}
\end{equation}

Notice from \eqref{eq:recover_parts} that $j = 0$ already corresponds to the original signal. Subbands that correspond to $j \in [1,L]$ are approximation parts of the signal, and we reconstruct them using only the approximation coefficients. Finally, subbands that correspond to $j \in [L+1,2L]$ are detail parts of signals, which are reconstructed using detail coefficients.

In the fMRI data acquisition experiments, total of $T = 1940$ measurements are obtained from a single participant. Since the  decomposition of a signal into more than 11 levels would not provide statistically sufficient information, we take $L = 11$. For each level, we decompose a region representative signal into $l + 1$ dyadic frequency subbands, and reconstruct $l + 1$ signals that contain a single approximation part $A_l$ and detail parts of that level and finer levels $\{ D_1, \ldots, D_l \}$. We use cubic Battle-Lemarie wavelets, which are frequently used in wavelet analysis of fMRI signals \cite{Richiardi2013,Richiardi2011}.


\subsection{Representation of Brain Connectivity by Mesh Networks}

In the second step of HMMNs, we estimate a brain network, called mesh network, to represent a cognitive task in terms of the interaction among anatomic regions, at each resolution. Multi-resolution networks enable us to analyze the topological properties of the brain, in different subbands. In addition, we employ graph embedding of the multi-resolution networks for extraction of features for decoding the cognitive tasks. An abstract representation of the brain states by  networks also helps us to visualize functional connectivity among the anatomic regions and describe their non-trivial properties in a compact way \cite{Richiardi2013}, \cite{Fallani2014}. During the analysis of brain topology,  efficiency metric is used to characterize the small-world properties of brain networks \cite{Achard2007,Bassett2006}. On the other hand, centrality metrics including betweenness and node degree are used to identify the significant anatomic regions \cite{Fallani2014}. In this work, we form a brain network for each decomposed subband of the fMRI signal obtained during a cognitive task with class label $c$, where $c = 1, 2, \ldots, 7$.

Multi-resolution network representation of fMRI data,  recorded during a task experiment session, is defined by the connectivity graph, {$G_{j,q}= \{V, A_{j,q}: \forall j \in \mathcal{J} \}$}, for each task experiment session, $q$, and for each subband, $j$.  In this graph, the set of vertices, $V$, corresponds to the ordered set of anatomic regions, $V = [r], \forall r \in R$. Vertex attributes are the average time series $\bm{x}_{j,q,r}$ obtained during a task experiment session, $q$, and for the $j^{th}$ subband.  The arc weights, $A_{j,q} = \{a_{j,q,r,s}\}_{r,s = 1} ^R$, between regions $r$ and $s$, for each task ($q$), are estimated from the local meshes of representative time series obtained at subband $j$, as will be described in the next section. 

\subsection{Formation of Local Meshes for Estimation of the Arc Weights of Multi-resolution Mesh Networks}

We form a mesh in a pre-defined neighborhood of each anatomic region. The nodes of the mesh correspond to the regions, which are represented by the average time series. Each node is connected to its $p$-nearest neighboring nodes to form a star mesh around an anatomic region. The neighboring regions (nodes) of each anatomic region are the regions with $p$-largest Pearson correlation with that region. The number of nearest neighbors, $p$, is determined by cross validation in the brain decoding phase.


For each mesh formed around an anatomic region,$r$, we estimate the arc weights for the $q^{th}$ task experiment session at the $j^{th}$ subband using the following regularized linear model;
\begin{equation}
	\label{eq:mesh}
	\bm{x}_{j,q,r} = \sum_{s \in \eta_p[r]}{ {a_{j,q,r,s}} \ \bm{x}_{j,q,s} + \lambda \  |a_{j,q,r,s}|^2  + \bm{\varepsilon}_{j,q,r}} ,
\end{equation}
where $\lambda$ is the regularization parameter. The mesh arc weights, $a_{j,q,r,s}$, are estimated by minimizing the error variance, $\bm{\varepsilon_{j,q,r}}$. In \eqref{eq:mesh},  $\bm{x_{j,q,r}}$ is a vector that represents the average voxel time series in region $r$ for the $j^{th}$ subband and for the $q^{th}$ task experiment session with  $D_q$ scans, such that,
\begin{equation}
\label{eq:interval}
\bm{x}_{j,q,r} = [x_{j,q,r}(1), x_{j,q,r}(2), \ldots , x_{j,q,r}(D_q)].
\end{equation}

The mesh defined in \eqref{eq:mesh} is solved for each region $r$ with its neighbors separately. In other words, we obtain an independent local mesh around each region, $r$. After we estimate all of the mesh arc weights, $A_{j,q} = \{a_{j,q,r,s}\}_{r,s = 1} ^R$, we represent the connectivity graph, {$G_{j,q}= \{V, A_{j,q}: \forall j \in \mathcal{J} \}$}, as an ensemble of all local meshes. Thus, we call {$G_{j,q}= \{V, A_{j,q}: \forall j \in \mathcal{J} \}$} as mesh network. Note that, mesh network is directed, since the relationship between two regions $r$ and $s$ differ based on the seed region. In other words, $ a_{j,q,r,s} \neq a_{j,q,s,r} $ since $r$ is the seed region used in the former representation, and $s$ is the seed region used in the latter one. Hence, the resulting graph is directed and asymmetric.

Mesh networks are estimated for both original fMRI signal, and its  approximation and detail parts of different resolutions. In other words, we form $2L+1$ distinct mesh networks for the frequency subbands $\{A_0, A_1, A_2, \ldots ,A_L, D_1, D_2, \ldots ,D_L\}$. Recall that, in our dataset, we use $L = 11$, and $l=0$ corresponds to the raw fMRI signal.

\subsection{Graph Embedding of Multi-resolution Mesh Networks}

To this end, we form a brain network for each task experiment, as ensemble of local meshes and estimate mesh arc weights at each subband. Then, we concatenate the estimated weights under a structured feature matrix to form the adjacency matrix of the mesh network at each subband.
 
For each task experiment, we form a set of multi-resolution brain networks whose arc weights remain the same during the experiment. Recall that, arc weights represent the relationship between seed region and its functionally nearest neighbors. If two regions $r$ and $s$ are not neighbors in task $q$, then $a_{j,q,r,s} = 0, \forall j$, for that task.

We form an adjacency matrix of size $R \times R$ for a mesh network using the estimated arc weights $(\forall_{r,s} a_{j,q,r,s})$. By concatenating the arc weights under a vector of size $1 \times R^2$, we embed the brain network for a task experiment $q$ at subband $j$, by the feature vector
$F_{j,q} = [a_{j,q,r,s}]_{r,s = 1}^R$. 

Note that, we compute a graph embedding for each task experiment obtained from a subject in the dataset. When we concatenate all feature vectors, we obtain a feature matrix $\mathcal{F}_{j} = [F_{j,q}]_{\forall q}$, for the $j^{th}$ subband. We assign a class label $c  \in [1,7] $ to each mesh network extracted from the task experiment $q$.

\subsection{Fusion of Multi-resolution Brain Networks  Under a Hierarchical Ensemble Learning Architecture}

One of the major problems of representing the cognitive tasks by multi-resolution brain networks is the complexity of the analysis and decoding. Furthermore, fusion of the information embedded in the multi-resolution networks requires the diversity of network structure and topology. In other words, the estimated arc weights in different resolutions should complement each other in some way, so that fusion process boosts the performance of the decoding task. In this study, we propose a fast and efficient ensemble learning method for fusion of the brain networks obtained in multiple resolution by adopting  a state-of-the-art ensemble learning method, called fuzzy stacked generalization (FSG), which was shown to perform better than various classifiers for MVPA \cite{Ozay2016}. We propose a two-phase approach  to employ the basic structure of FSG in our proposed framework as follows (see Figure~\ref{fig:flowchart}):


First, we train an individual logistic regression classifier for each subband $j$ at the base-layer of an FSG. The dataset is partitioned into {$k$-folds}, where  $k-1$ of them is used for training by one-leave out cross validation and the $k^{th}$ fold is used to test the classifiers. Then, we validate classification (decision) hypothesis of each classifier using the samples belonging to the $k^{th}$ chunk of the data. Each classifier outputs the estimate of the posterior probabilities of a mesh network belonging to a particular class. We put the posterior probabilities in a  membership vector of size $1 \times C$ for all tasks, where $C$ is the number of classes. Thereby, each base layer classifier maps $R^2=8100$ dimensional mesh arc feature vectors obtained from each subband to a $C=7$ dimensional decision space. This mapping trains the base layer classifiers to become experts  for capturing diverse connectivity properties of fMRI data signified in different subbands. 

Next, we compute class membership vectors of test samples using decisions of the trained classifiers for test data. At the meta-layer, we obtain a membership vector of size $1 \times CE$ for each sample, by concatenating membership vectors obtained from all base-layer classifiers, where $E$ is the number of classifiers (i.e., number of the subbands used to train base-layer classifiers). 

Then, we train and test meta-layer classifier using the concatenated membership vectors of the training and test data, respectively. In the experiments, we provide classification results using logistic regression and Support Vector Machines (SVM) classifiers with linear kernels at the meta-layer. In the experiments, $E$ is varied by selecting various combinations of subbands to see the effect of approximation and detail parts on the classification performance. 

\renewcommand\theadfont{\bfseries}

\begin{table*}[t]
\centering
\caption{Classification performances (\%) of approximation and detail parts of decomposition levels $l \in [0,11]$ at the output of base-layer classifiers of FSG, trained and tested by three sets of featue vectors, namely, mesh arc weights, pairwise correlation and region representative time series.}
\label{tab:single-subband}
\begin{tabular}{ccccccccc}
\hline
                 & \multicolumn{2}{c}{\thead{\textbf{Mesh Arc} \\\textbf{ Weights}}}  &    & \multicolumn{2}{c}{\thead{\textbf{Pairwise} \\ \textbf{Correlations}}} & & \multicolumn{2}{c}{\thead{\textbf{Representative}\\ \textbf{Time Series}}}               \\ \hline
            \textbf{Level}     & \textbf{$A_l$} & \textbf{$D_l$} & & \textbf{$A_l$} & \textbf{$D_l$} & & \textbf{$A_l$} & \textbf{$D_l$} \\ 
            \cmidrule[.8pt]{2-3} 	
            \cmidrule[.8pt]{5-6} 
            \cmidrule[.8pt]{8-9} 

\textbf{0}  & 97.15 & -     && 89.97 & -     && 16.11 & -     \\
\textbf{1}  & 97.33 & 44.20 && \textbf{90.38} & 33.73 && 16.14 & 38.97 \\
\textbf{2}  & \textbf{97.42} & 52.19 && 90.31 & 42.24 && 16.09 & 38.19 \\
\textbf{3}  & 97.28 & 82.81 && 89.59 & 76.11 && 16.25 & 37.31 \\
\textbf{4}  & 97.01 & 92.49 && 88.22 & 83.04 && 16.23 & 49.73 \\
\textbf{5}  & 94.54 & \textbf{95.99} && 81.49 & \textbf{90.89} && 16.35 & 60.84 \\
\textbf{6}  & 70.65 & 93.37 && 48.76 & 85.43 && \textbf{16.57} & \textbf{69.84} \\
\textbf{7}  & 42.36 & 62.39 && 34.53 & 49.31 && 15.98 & 56.90 \\
\textbf{8}  & 25.02 & 35.29 && 25.95 & 32.41 && 15.56 & 34.48 \\
\textbf{9}  & 14.29 & 14.29 && 21.57 & 28.62 && 16.14 & 34.07 \\
\textbf{10} & 15.59 & 14.29 && 16.69 & 22.21 && 15.01 & 45.54 \\
\textbf{11} & 14.29 & 14.29 && 16.71 & 25.88 && 14.29 & 34.81 \\
\hline
\end{tabular}
\end{table*}




Note that, fusion of the collaboratively trained expert classifiers provides higher performance than the majority voting methods \cite{Ozay2016}. In the experimental analyses, we examined this observation using the proposed framework, and compared the results with state-of-the-art majority voting methods for fMRI data analysis \cite{Richiardi2011}. 

\section{Decoding Performances and Topology Analysis of Hierarchical 
Multi-Resolution Mesh Networks}
In this section, first, we test the representation power of  HMMNs  for brain decoding, in different subbands. Then, we test the boosting effect of fusion of all subbands. After that, we analyze the network topology of HMMNs in multiple subbands, in terms of the metrics, such as node degree, node strength, betweenness centrality and global efficiency. Finally, we investigate the diversity properties of HMMNs at base-layer classifiers. The represention power of HMMNs is tested on the task fMRI dataset of HCP. 
\subsection{Brain Decoding by Hierarchical Multi-resolution Mesh Networks}


In order to decode the cognitive tasks of HCP dataset, first, we decompose the original fMRI signals into $L=11$ levels, and obtain $2L+1$ subbands to compute multi-resolution representations of fMRI data. Next, for each subband, we estimate the mesh arc weights, at each resolution to obtain total of 2L+1=23 coarse-to-fine brain networks for each task.

\begin{table*}[t!]
\centering
\caption{Classification performances (\%) of fuzzy stacked generalization, majority voting (MV) and weighted majority voting (WMV) methods. At meta-layer of FSG two types of classifiers are used, namely, logistic regression (FSG-L) and Support Vector Machines (FSG-S. Numbers with + or - sign show the standard deviation of the performances obtained during the cross validation.}
\label{tab:wavelet-overlap}
\begin{tabular}{lccccc}
\hline
                                                                                   &                                         \textbf{ Levels}     & \textbf{FSG-L} & \textbf{FSG-S} & \textbf{MV}    & \textbf{WMV}   \\ \hline
\multirow{10}{*}{\textbf{Mesh Arc Weights}}                                                             & \textbf{2}  & 97.83 $\pm$ 0.82 & 97.97 $\pm$ 0.75 & 97.47 $\pm$ 0.78 & 98.51 $\pm$ 0.73 \\
 & \textbf{3}  & 98.13 $\pm$ 0.72 & 98.29 $\pm$ 0.67 & 97.88 $\pm$ 0.94 & 98.3 $\pm$ 0.71  \\
 & \textbf{4}  & 98.55 $\pm$ 0.75 & 98.69 $\pm$ 0.71 & 98.07 $\pm$ 1.05 & 98.3 $\pm$ 0.81  \\
 & \textbf{5}  & 98.8 $\pm$ 0.74  & 99.1 $\pm$ 0.58  & 98.5 $\pm$ 0.85  & 98.37 $\pm$ 0.81 \\
 & \textbf{6}  & 98.97 $\pm$ 0.49 & 99.22 $\pm$ 0.44 & 98.57 $\pm$ 0.77 & 98.46 $\pm$ 0.82 \\
 & \textbf{7}  & 99.05 $\pm$ 0.54 & 99.4 $\pm$ 0.46  & 98.62 $\pm$ 0.74 & 98.53 $\pm$ 0.8  \\
 & \textbf{8}  & 99.05 $\pm$ 0.51 & 99.49 $\pm$ 0.43 & \textbf{98.8} $\pm$ 0.65  & 98.55 $\pm$ 0.82 \\
 & \textbf{9}  & 99.13 $\pm$ 0.56 & 99.63 $\pm$ 0.4  & \textbf{98.8} $\pm$ 0.6   & \textbf{98.6 }$\pm$ 0.75  \\
 & \textbf{10} & 99.15 $\pm$ 0.56 & 99.61 $\pm$ 0.46 & 98.71 $\pm$ 0.57 & \textbf{98.6 }$\pm$ 0.75  \\
 & \textbf{11} & \textbf{99.15} $\pm$ 0.57 & \textbf{99.63 }$\pm$ 0.45 & 98.25 $\pm$ 0.71 & \textbf{98.6} $\pm$ 0.75  
\\ \hline
\multirow{10}{*}{\textbf{\begin{tabular}[c]{@{}l@{}}Pairwise \\ Correlation\end{tabular}}}  & \textbf{2}  & 91.21 $\pm$ 2.03 & 91.23 $\pm$ 1.92 & 90.51 $\pm$ 2.25 & \textbf{94.86} $\pm$ 1.28 \\
 & \textbf{3}  & 92.36 $\pm$ 1.73 & 92.36 $\pm$ 1.92 & 90.75 $\pm$ 1.88 & 92.45 $\pm$ 1.87 \\
 & \textbf{4}  & 93.9 $\pm$ 1.02  & 93.88 $\pm$ 1.19 & 91.46 $\pm$ 1.73 & 91.94 $\pm$ 1.71 \\
 & \textbf{5}  & 95.67 $\pm$ 1.16 & 95.49 $\pm$ 1.31 & 92.24 $\pm$ 1.39 & 91.92 $\pm$ 1.7  \\
 & \textbf{6}  & 96.41 $\pm$ 1.11 & 96.32 $\pm$ 1.27 & 93.02 $\pm$ 1.28 & 92.1 $\pm$ 1.55  \\
 & \textbf{7}  & 96.66 $\pm$ 1.26 & 96.46 $\pm$ 1.22 & 93.26 $\pm$ 1.34 & 92.27 $\pm$ 1.46 \\
 & \textbf{8}  & \textbf{96.85} $\pm$ 1.16 & 96.45 $\pm$ 1.28 & 93.32 $\pm$ 1.4  & 92.34 $\pm$ 1.5  \\
 & \textbf{9}  & 96.84 $\pm$ 1.17 & \textbf{96.64} $\pm$ 1.04 & 93.55 $\pm$ 1.31 & 92.42 $\pm$ 1.57 \\
 & \textbf{10} & 96.78 $\pm$ 1.18 & 96.41 $\pm$ 1.17 & 93.42 $\pm$ 1.26 & 92.47 $\pm$ 1.57 \\
 & \textbf{11} & 96.87 $\pm$ 1.26 & 96.41 $\pm$ 1.38 & \textbf{93.58} $\pm$ 1.18 & 92.54 $\pm$ 1.6  \\ \hline
\multirow{10}{*}{\textbf{\begin{tabular}[c]{@{}l@{}}Representative \\ Time  Series\end{tabular}}}                                                        
 & \textbf{2}  & 38.38 $\pm$ 1.92 & 37.41 $\pm$ 2.22 & 16.2 $\pm$ 1.51  & \textbf{80.99} $\pm$ 1.73 \\
 & \textbf{3}  & 47.03 $\pm$ 2.29 & 47.17 $\pm$ 2.27 & 16.25 $\pm$ 1.35 & 72.07 $\pm$ 1.82 \\
 & \textbf{4}  & 64.76 $\pm$ 3.04 & 66.16 $\pm$ 2.82 & 16.67 $\pm$ 1.48 & 64.76 $\pm$ 1.86 \\
 & \textbf{5}  & 78.87 $\pm$ 1.55 & 80.69 $\pm$ 1.4  & 17.86 $\pm$ 2.06 & 62.52 $\pm$ 1.78 \\
 & \textbf{6}  & 90.21 $\pm$ 1.41 & 91.88 $\pm$ 1.21 & 19.36 $\pm$ 3.05 & 64.11 $\pm$ 2.5  \\
 & \textbf{7}  & 92.54 $\pm$ 1.71 & 93.53 $\pm$ 1.23 & 21.38 $\pm$ 3.68 & 66.76 $\pm$ 2.78 \\
 & \textbf{8}  & 92.95 $\pm$ 1.64 & 94.01 $\pm$ 1.51 & 23.71 $\pm$ 3.59 & 68.25 $\pm$ 2.53 \\
 & \textbf{9}  & 93.39 $\pm$ 1.51 & 94.57 $\pm$ 1.1  & 27.46 $\pm$ 3.97 & 69.68 $\pm$ 2.56 \\
 & \textbf{10} & 94.89 $\pm$ 1.02 & 96.22 $\pm$ 1    & 31.21 $\pm$ 3.32 & 71.46 $\pm$ 2.4  \\
 & \textbf{11} & \textbf{95.72} $\pm$ 1.18 & \textbf{96.64} $\pm$ 1.16 & \textbf{32.41} $\pm$ 2.81 & 72.07 $\pm$ 2.38 \\ \hline
\end{tabular}
\end{table*}

In order to optimize the mesh size $p$ and the regularization parameter $\lambda$, we perform cross validation on the mesh arc weights of the original signal. We searched for the optimal parameters for $p \in \{5,10, \ldots , 40\}$ and $\lambda \in \{8,16,32, \ldots, 512\}$. Our cross validation results show that $p = 40$ and $\lambda = 32$ are the optimal values. In the following sections, we use these values while estimating the mesh weights for all subbands.

For comparison, we also compute the connectivity graphs where arc weights correspond to Pearson correlation coefficients computed between pairs of regions. Finally, as a baseline, we train and test the classifiers using vectors of representative time series, $x_{j,c,r}$, which are computed for each anatomic region and for each subband. Thus, we train logistic regression classifiers by three feature sets, namely, vector of mesh arc weights, pairwise correlations and vector of representative time series. Then, we compute classification performance for each subband, separately. 

Table~\ref{tab:single-subband} shows classification  accuracy obtained for features extracted using approximation and detail parts of various levels of decomposition. Note that $A_0$ corresponds to the original fMRI signal.

It is observed that feature vectors formed by mesh arc weights perform substantially better compared to Pearson's pairwise correlation and the region representative time series for each subband. We observe a decrease in the classification performance of approximation part as the decomposition level increases, decaying to the chance performance at $11^{th}$ level. On the contrary, classification performance of detail parts increases up to $l = 5$, and then decreases up to level $l = 11$. For $l \in [5,8]$, the information  content  obtained in the detail part (high resolution part) is more discriminative compared to approximation part (low resolution part).

\begin{figure*}[t]
	\centering
    \includegraphics[width=.9\textwidth]{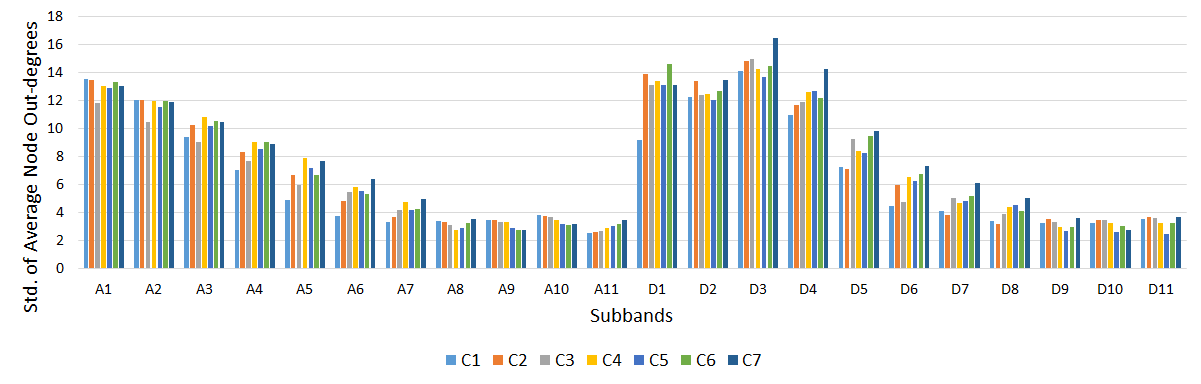}			
	\caption{Standard deviation of average node out-degrees.}
    \label{fig:node_outdegree}
\end{figure*}

\captionsetup[subfigure]{skip=0pt} 
\begin{figure*}[t!]
	\captionsetup[subfigure]{labelformat=empty}	
	\centering
	\begin{subfigure}[t]{0.16\textwidth}
		\includegraphics[width=\textwidth]{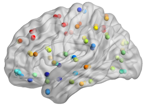}
		\caption{$A_1$}
	\end{subfigure}	\hfill
    \begin{subfigure}[t]{0.16\textwidth}
		\includegraphics[width=\textwidth]{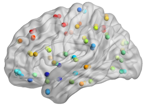}
		\caption{$A_2$} 
	\end{subfigure}\hfill
    \begin{subfigure}[t]{0.16\textwidth}
		\includegraphics[width=\textwidth]{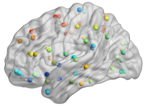}
		\caption{$A_3$}
	\end{subfigure}\hfill
    \begin{subfigure}[t]{0.16\textwidth}
		\includegraphics[width=\textwidth]{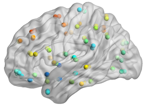}
		\caption{$A_4$}
	\end{subfigure}	\hfill
    \begin{subfigure}[t]{0.16\textwidth}
		\includegraphics[width=\textwidth]{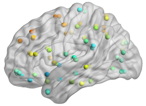}
		\caption{$A_5$}
	\end{subfigure}	\hfill  	
	\begin{subfigure}[t]{0.16\textwidth}
		\includegraphics[width=\textwidth]{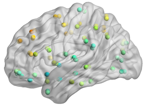}
		\caption{$A_6$}
	\end{subfigure} \hfill
    \begin{subfigure}[t]{0.16\textwidth}
		\includegraphics[width=\textwidth]{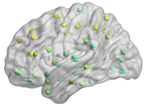}
		\caption{$A_7$}
	\end{subfigure}	\hfill
    \begin{subfigure}[t]{0.16\textwidth}
		\includegraphics[width=\textwidth]{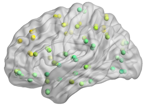}
		\caption{$A_8$} 
	\end{subfigure}\hfill
    \begin{subfigure}[t]{0.16\textwidth}
		\includegraphics[width=\textwidth]{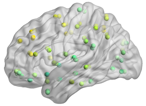}
		\caption{$A_9$}
	\end{subfigure}\hfill
    \begin{subfigure}[t]{0.16\textwidth}
		\includegraphics[width=\textwidth]{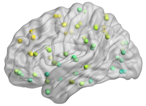}
		\caption{$A_{10}$}
	\end{subfigure}	\hfill
    \begin{subfigure}[t]{0.16\textwidth}
		\includegraphics[width=\textwidth]{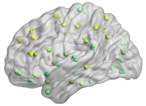}
		\caption{$A_{11}$}
	\end{subfigure}	\hfill  	
	\begin{subfigure}[t]{0.16\textwidth}
		\includegraphics[trim={5cm 1cm 4cm 1cm},clip, width=\textwidth]{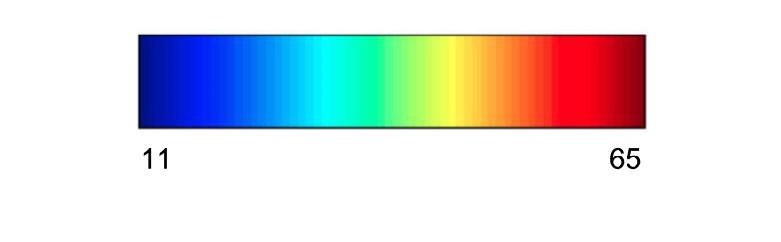}
	\end{subfigure}

    	\begin{subfigure}[t]{0.16\textwidth}
		\includegraphics[width=\textwidth]{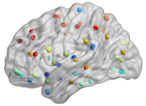}
		\caption{$D_1$}
	\end{subfigure}	\hfill
    \begin{subfigure}[t]{0.16\textwidth}
		\includegraphics[width=\textwidth]{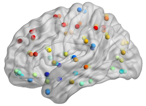}
		\caption{$D_2$} 
	\end{subfigure}\hfill
    \begin{subfigure}[t]{0.16\textwidth}
		\includegraphics[width=\textwidth]{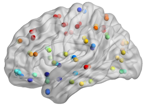}
		\caption{$D_3$}
	\end{subfigure}\hfill
    \begin{subfigure}[t]{0.16\textwidth}
		\includegraphics[width=\textwidth]{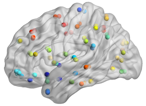}
		\caption{$D_4$}
	\end{subfigure}	\hfill
    \begin{subfigure}[t]{0.16\textwidth}
		\includegraphics[width=\textwidth]{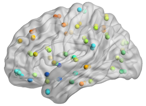}
		\caption{$D_5$}
	\end{subfigure}	\hfill  	
	\begin{subfigure}[t]{0.16\textwidth}
		\includegraphics[width=\textwidth]{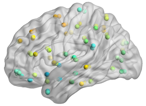}
		\caption{$D_6$}
	\end{subfigure} \hfill
    	\begin{subfigure}[t]{0.16\textwidth}
		\includegraphics[width=\textwidth]{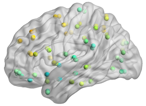}
		\caption{$D_7$}
	\end{subfigure}	\hfill
    \begin{subfigure}[t]{0.16\textwidth}
		\includegraphics[width=\textwidth]{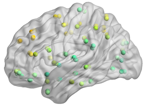}
		\caption{$D_8$} 
	\end{subfigure}\hfill
    \begin{subfigure}[t]{0.16\textwidth}
		\includegraphics[width=\textwidth]{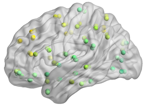}
		\caption{$D_9$}
	\end{subfigure}\hfill
    \begin{subfigure}[t]{0.16\textwidth}
		\includegraphics[width=\textwidth]{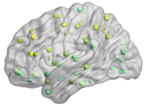}
		\caption{$D_{10}$}
	\end{subfigure}	\hfill
    \begin{subfigure}[t]{0.16\textwidth}
		\includegraphics[width=\textwidth]{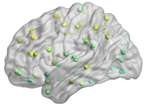}
		\caption{$D_{11}$}
	\end{subfigure}	\hfill  	
	\begin{subfigure}[t]{0.16\textwidth}
		\includegraphics[trim={5cm 1cm 4cm 1cm},clip, width=\textwidth]{colorbar_degree.jpg}
        \caption{}
	\end{subfigure}
    \vspace{0.4cm}
	\caption{Overlay of average node out-degrees on a human brain template.}
    \label{fig:brain_degree}
\end{figure*}

\begin{figure*}[t!]
	\centering
    \includegraphics[trim={5cm 0cm 4.2cm 1cm},clip, width=.85\textwidth]{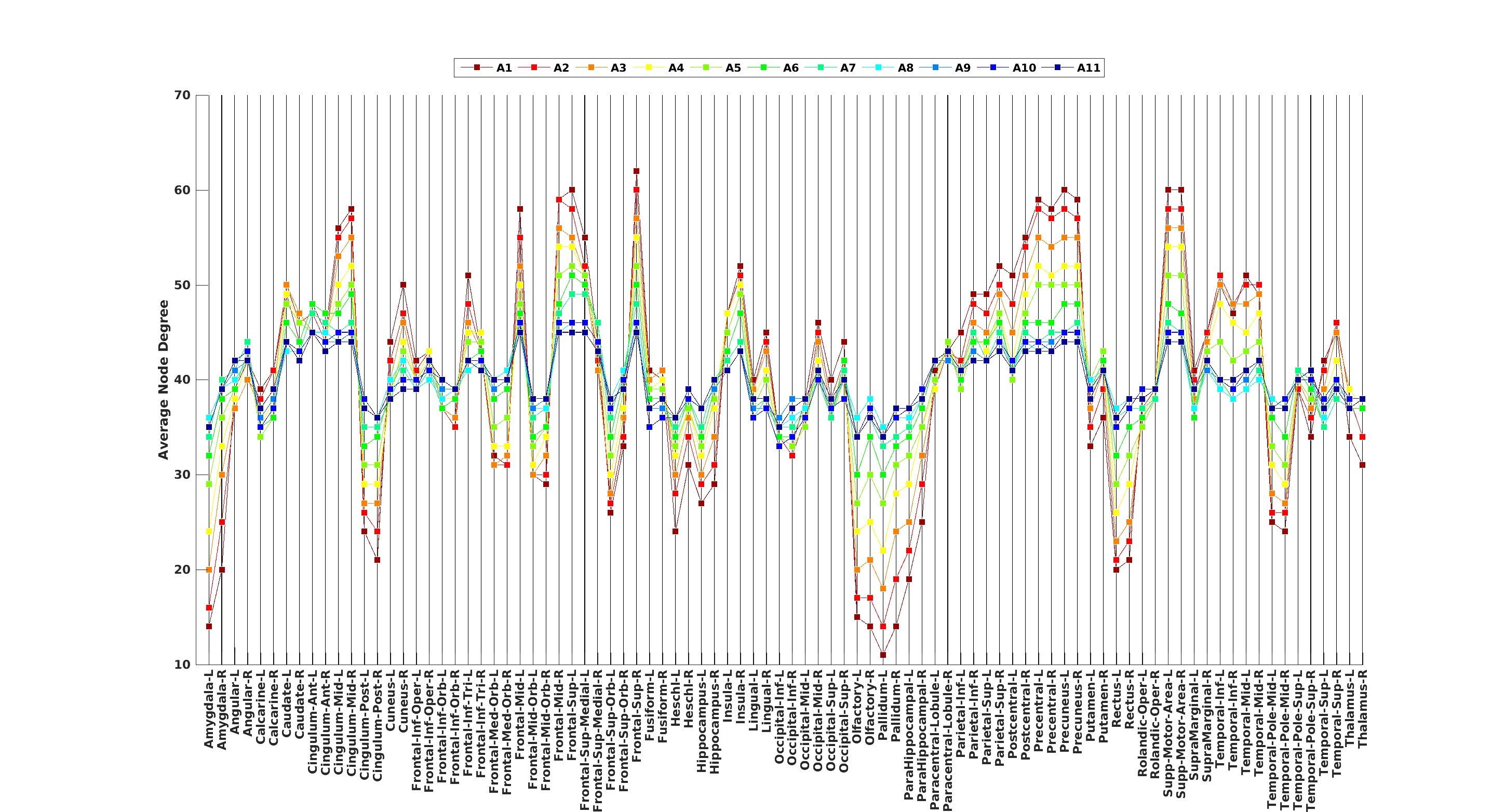}			
	\caption{Average node out-degrees obtained for subbands $[A_1, A_{11}]$ for anatomical regions. Maximum values show the massively connected anatomic regions whereas the minimum values show the rarely connected regions.}
    \label{fig:node_degree_point_app}
\end{figure*}

\begin{figure*}[t!]
	\centering
    \includegraphics[trim={5cm 0cm 4.2cm 1cm},clip, width=.85\textwidth]{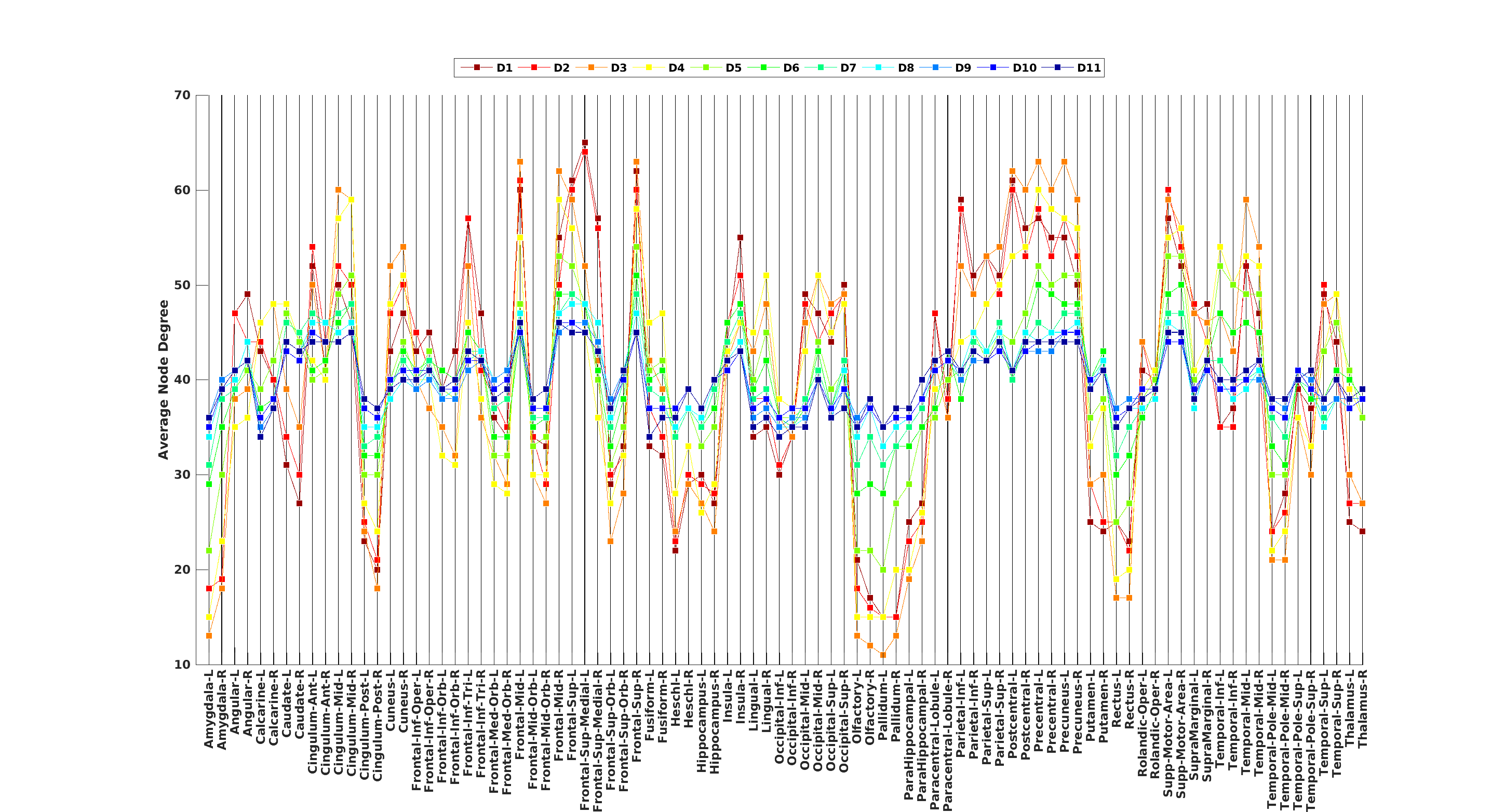}			
	\caption{Average node out-degrees obtained for subbands $[D_1, D_{11}]$ for anatomical regions. Maximum and minimum values which show the massively and rarely connected regions meet with that of the approximation parts of figure\ref{fig:node_degree_point_app}}
    \label{fig:node_degree_point_det}
\end{figure*}

\captionsetup[subfigure]{skip=0pt} 
\begin{figure*}[t]
	\captionsetup[subfigure]{labelformat=empty}	
	\centering
	\begin{subfigure}[t]{0.16\textwidth}
		\includegraphics[width=\textwidth]{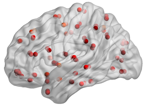}
		\caption{$A_1$}
	\end{subfigure}	\hfill
    \begin{subfigure}[t]{0.16\textwidth}
		\includegraphics[width=\textwidth]{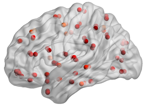}
		\caption{$A_2$} 
	\end{subfigure}\hfill
    \begin{subfigure}[t]{0.16\textwidth}
		\includegraphics[width=\textwidth]{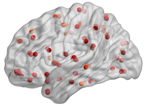}
		\caption{$A_3$}
	\end{subfigure}\hfill
    \begin{subfigure}[t]{0.16\textwidth}
		\includegraphics[width=\textwidth]{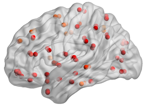}
		\caption{$A_4$}
	\end{subfigure}	\hfill
    \begin{subfigure}[t]{0.16\textwidth}
		\includegraphics[width=\textwidth]{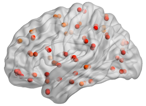}
		\caption{$A_5$}
	\end{subfigure}	\hfill  	
	\begin{subfigure}[t]{0.16\textwidth}
		\includegraphics[width=\textwidth]{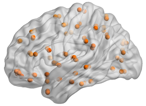}
		\caption{$A_6$}
	\end{subfigure} \hfill
    \begin{subfigure}[t]{0.16\textwidth}
		\includegraphics[width=\textwidth]{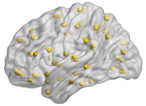}
		\caption{$A_7$}
	\end{subfigure}	\hfill
    \begin{subfigure}[t]{0.16\textwidth}
		\includegraphics[width=\textwidth]{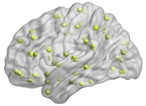}
		\caption{$A_8$} 
	\end{subfigure}\hfill
    \begin{subfigure}[t]{0.16\textwidth}
		\includegraphics[width=\textwidth]{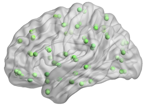}
		\caption{$A_9$}
	\end{subfigure}\hfill
    \begin{subfigure}[t]{0.16\textwidth}
		\includegraphics[width=\textwidth]{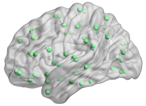}
		\caption{$A_{10}$}
	\end{subfigure}	\hfill
    \begin{subfigure}[t]{0.16\textwidth}
		\includegraphics[width=\textwidth]{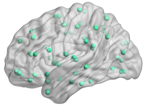}
		\caption{$A_{11}$}
	\end{subfigure}	\hfill  	
	\begin{subfigure}[t]{0.16\textwidth}
		\includegraphics[trim={5cm 1cm 4cm 1cm},clip, width=\textwidth]{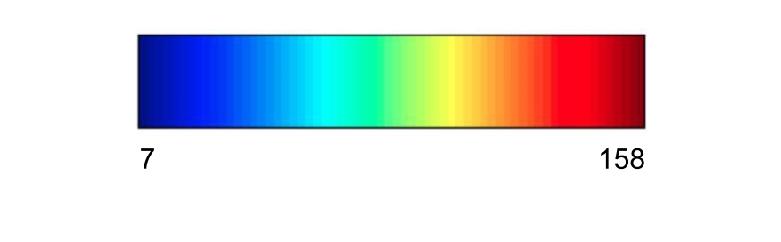}
	\end{subfigure}

    	\begin{subfigure}[t]{0.16\textwidth}
		\includegraphics[width=\textwidth]{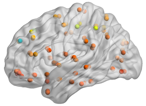}
		\caption{$D_1$}
	\end{subfigure}	\hfill
    \begin{subfigure}[t]{0.16\textwidth}
		\includegraphics[width=\textwidth]{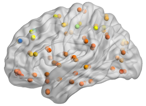}
		\caption{$D_2$} 
	\end{subfigure}\hfill
    \begin{subfigure}[t]{0.16\textwidth}
		\includegraphics[width=\textwidth]{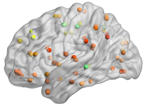}
		\caption{$D_3$}
	\end{subfigure}\hfill
    \begin{subfigure}[t]{0.16\textwidth}
		\includegraphics[width=\textwidth]{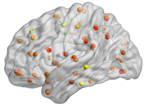}
		\caption{$D_4$}
	\end{subfigure}	\hfill
    \begin{subfigure}[t]{0.16\textwidth}
		\includegraphics[width=\textwidth]{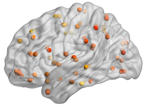}
		\caption{$D_5$}
	\end{subfigure}	\hfill  	
	\begin{subfigure}[t]{0.16\textwidth}
		\includegraphics[width=\textwidth]{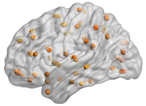}
		\caption{$D_6$}
	\end{subfigure} \hfill
    	\begin{subfigure}[t]{0.16\textwidth}
		\includegraphics[width=\textwidth]{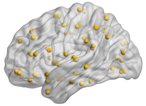}
		\caption{$D_7$}
	\end{subfigure}	\hfill
    \begin{subfigure}[t]{0.16\textwidth}
		\includegraphics[width=\textwidth]{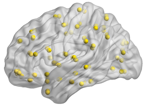}
		\caption{$D_8$} 
	\end{subfigure}\hfill
    \begin{subfigure}[t]{0.16\textwidth}
		\includegraphics[width=\textwidth]{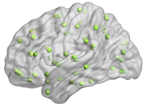}
		\caption{$D_9$}
	\end{subfigure}\hfill
    \begin{subfigure}[t]{0.16\textwidth}
		\includegraphics[width=\textwidth]{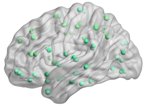}
		\caption{$D_{10}$}
	\end{subfigure}	\hfill
    \begin{subfigure}[t]{0.16\textwidth}
		\includegraphics[width=\textwidth]{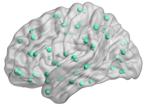}
		\caption{$D_{11}$}
	\end{subfigure}	\hfill  	
	\begin{subfigure}[t]{0.16\textwidth}
		\includegraphics[trim={5cm 1cm 4cm 1cm},clip, width=\textwidth]{colorbar_strength.jpg}
    
    \caption{}
	\end{subfigure}
    \vspace{0.4cm}
	\caption{Overlay of average node out-strengths on a human brain template.}
    \label{fig:brain_strength}
\end{figure*}

The posterior probabilities obtained at the output of the classifiers are fused to form the decision space by concatenating them under the same vector space. The vectors in the decision space are then used as the input to the meta-layer of FSG,  where both logistic regression and Support Vector Machine (SVM) classifiers are trained and tested. The results are provided in columns \textbf{FSG-L} and \textbf{FSG-S} of Table~\ref{tab:wavelet-overlap}. For comparison, we also fuse decisions of classifiers using majority voting (\textbf{MV}) and weighted majority voting (\textbf{WMV}) approaches. In majority voting method, all of the classifiers have equal confidence, whereas in weighted majority voting, the  classification performances of each classifier obtained in training step are used as weights. 

When the signal is decomposed into $L$ levels, we obtain total of $2L+1=23$ brain networks, for the original signal, $A_0$, for the approximation parts, $\{A_1, A_2, \ldots, A_L\}$ and the detail parts $\{D_1, D_2, \ldots, D_L\}$. The significance analysis of Section 4.3  reveals that subband $D_1$ does not provide additional information for any of the classes. Therefore, we discard $D_1$ and train total of 22 classifiers by the mesh networks extracted from each subband. 

Table~\ref{tab:wavelet-overlap} shows the performance of FSG formed by various sizes from 2 to 2xL base-layer classifiers using mesh arcs, Pairwise correlation and average region time series. Among them,  mesh arc weights give substantially better brain decoding performance at the output of FSG. We observe that the performances increase as the number of subbands increases when mesh arc weights or average voxel time series of regions are fed at the input of the base layer classifiers. However, the performances decrease after the $8^{th}$ level for the  pairwise correlation features, showing that they do not provide additional information after this resolution.

\begin{figure*}[t!]
	\centering
    \includegraphics[trim={4.8cm 0cm 4cm 1cm},clip, width=.85\textwidth]{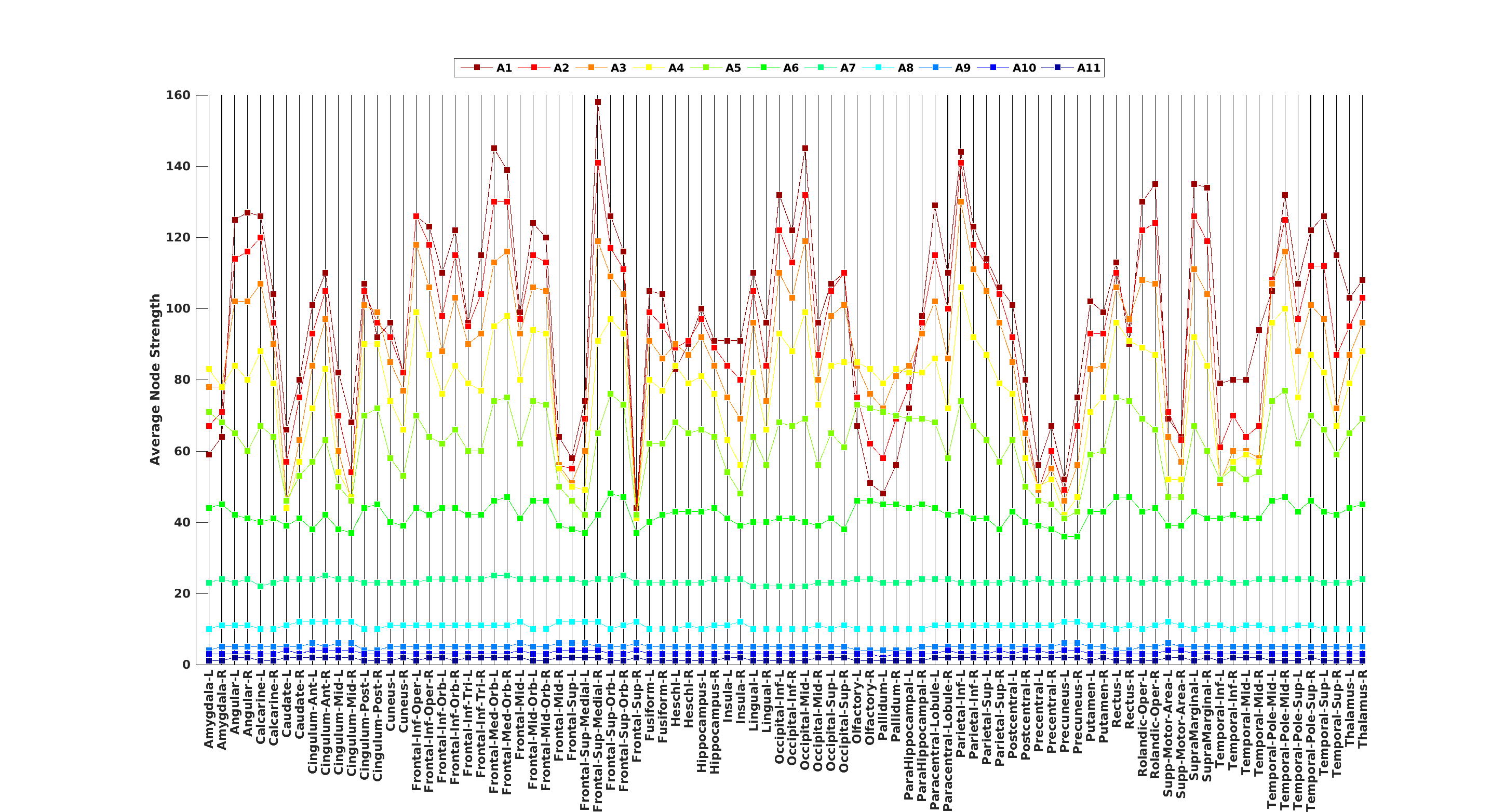}			
	\caption{Average node out-strength obtained for subbands $[A_1, A_{11}]$ for anatomical regions. Note that the strength and variations among the anatomic regions decreases as the level increases.
    }
    \label{fig:node_strength_point_app}
\end{figure*}

\begin{figure*}[t]
	\centering
    \includegraphics[trim={4.8cm 0cm 4cm 1cm},clip, width=.85\textwidth]{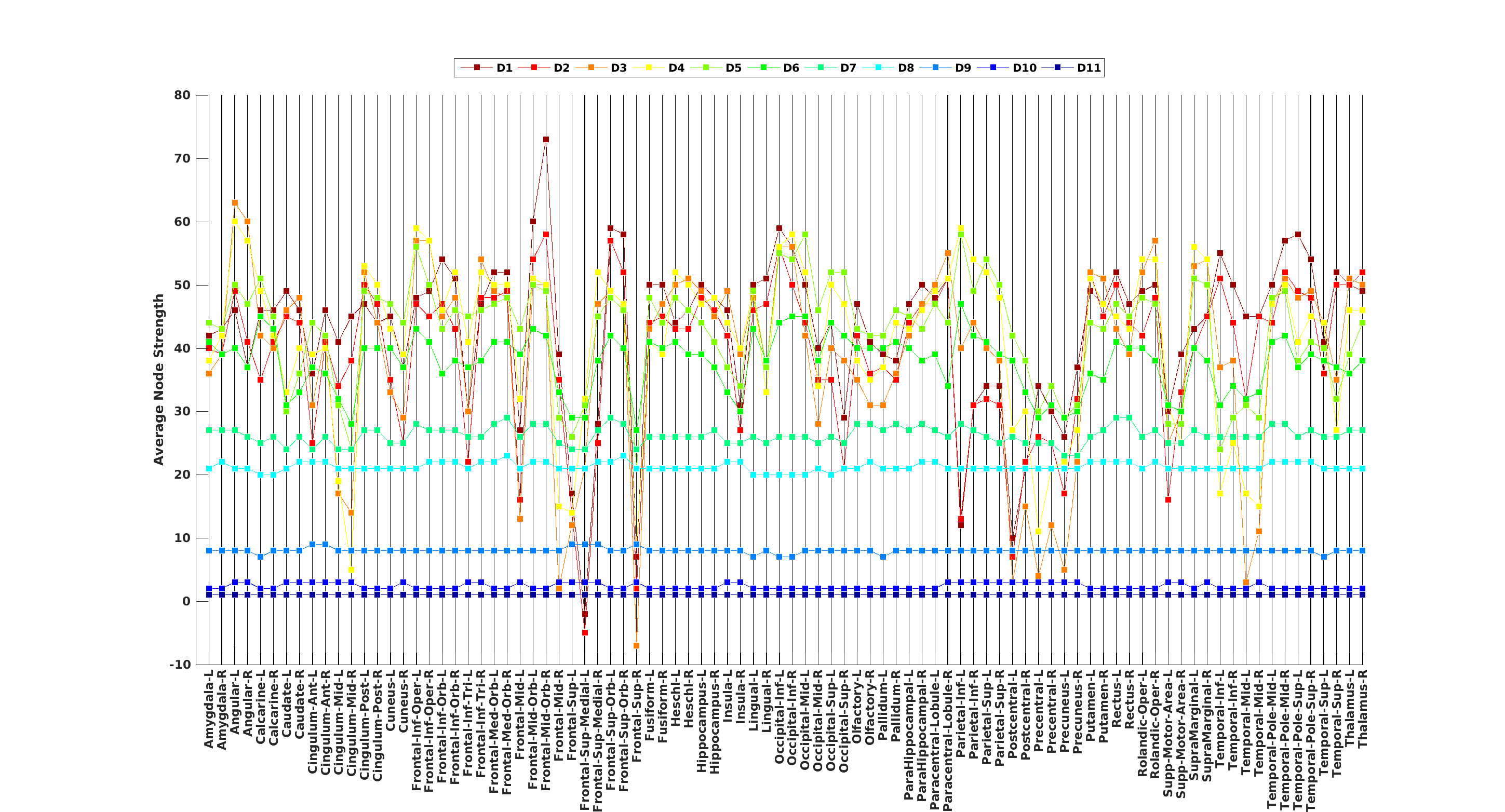}			
	\caption{Average node strength obtained for subbands $[D_1, D_{11}]$ for anatomical regions. Note that behavior of node strengths are similar to that of  approximation parts shown in Figure\ref{fig:node_strength_point_app}.}
    \label{fig:node_strength_point_det}
\end{figure*}


\subsection{Network Topology and Connectivity Analysis of Hierarchical Multi-resolution Mesh Networks}

In this sub-section, we analyze the network properties of the proposed  HMMNs, in different resolutions. Recall that the brain networks defined by mesh ensembles are weighted and directed. We compute graph connectivity measures, namely, node degree, node strength, node betweenness centrality and global efficiency using Brain Connectivity Toolbox \cite{Rubinov2010}. The figures of this section provide the average values of these measures over all subjects. 

\subsubsection{Connectivity Analysis by Node Degree}

Node degree represents the number of links connected to a node. While in-degree is the number of inward links, out-degree is the number of outward links. In this study, we select the number of nearest neighbors as $p = 40$, i.e. in-degree equals to $40$ for each node. Yet, we wish to explore the nodes (anatomic regions) in a mesh network, which are connected to many nodes through the outgoing mesh arc weights. In order to analyze this property, we compute the out-degree distribution of nodes. We also explore the variations of the node degrees with respect to subband. For this purpose, we measure the standard deviation of node out-degrees for all the tasks at each subband. Figure \ref{fig:node_outdegree} shows that the standard deviation of node out-degrees do not differ significantly over the  task classes at a subband. However, they vary greatly with respect to the subbands. As the decomposition level $l$ increases, the standard deviations of node out-degrees decreases, indicating that the regions become more and more similar with respect to node out degrees. On the other hand, for the detail parts, the standard deviation of node out-degrees slightly increase up to level, $l = 3$, then, fall down as the level increases to $l = 11$. This figure is consistent with the performance results of Table~\ref{tab:single-subband}, where we observe a monotonic decay in the performances of HMMNs as the level of the approximation part increases. Similarly, we notice that standard deviation of node out-degrees in the detail parts varies proportional to the decoding performances of Table~\ref{tab:single-subband} with respect to the levels. 

\begin{figure*}[htbp]
	\centering
	\begin{subfigure}[t]{.4\textwidth}
		\includegraphics[width=\textwidth]{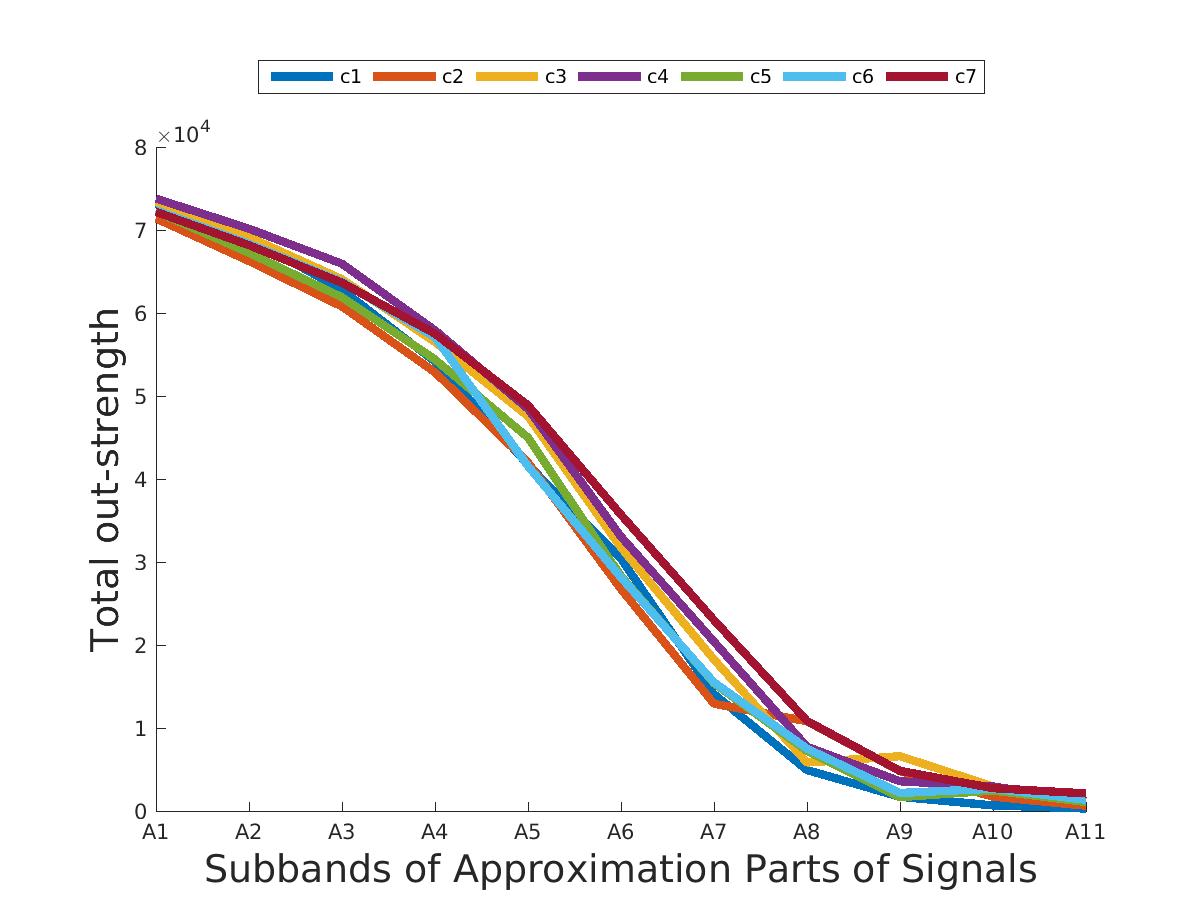}
		\label{fig:total_strength_app}
	\end{subfigure}%
    \begin{subfigure}[t]{.4\textwidth}
		\includegraphics[width=\textwidth]{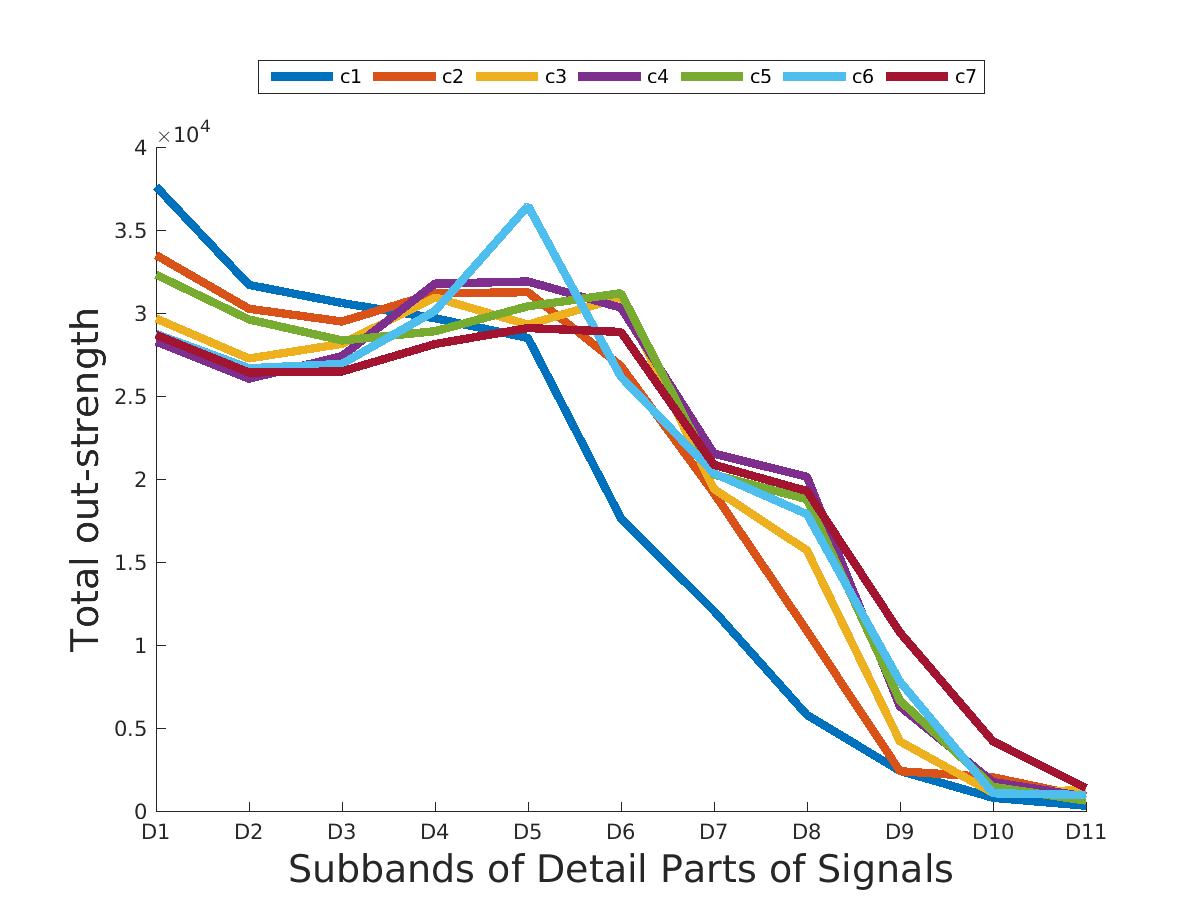}
		\label{fig:total_strength_det}
	\end{subfigure}
	\caption{Average total strength of all mesh networks formed for each task category, over the subbands, for approximation and detail parts.}
    \label{fig:total_strength}
\end{figure*}

\captionsetup[subfigure]{skip=0pt} 
\begin{figure*}[htbp]
	\captionsetup[subfigure]{labelformat=empty}	
	\centering
	\begin{subfigure}[t]{0.16\textwidth}
		\includegraphics[width=\textwidth]{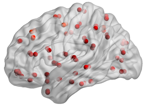}
		\caption{$A_1$}
	\end{subfigure}	\hfill
    \begin{subfigure}[t]{0.16\textwidth}
		\includegraphics[width=\textwidth]{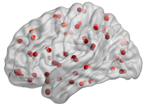}
		\caption{$A_2$} 
	\end{subfigure}\hfill
    \begin{subfigure}[t]{0.16\textwidth}
		\includegraphics[width=\textwidth]{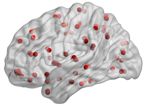}
		\caption{$A_3$}
	\end{subfigure}\hfill
    \begin{subfigure}[t]{0.16\textwidth}
		\includegraphics[width=\textwidth]{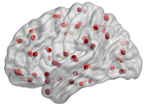}
		\caption{$A_4$}
	\end{subfigure}	\hfill
    \begin{subfigure}[t]{0.16\textwidth}
		\includegraphics[width=\textwidth]{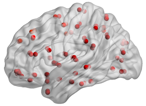}
		\caption{$A_5$}
	\end{subfigure}	\hfill  	
	\begin{subfigure}[t]{0.16\textwidth}
		\includegraphics[width=\textwidth]{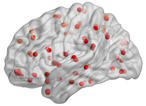}
		\caption{$A_6$}
	\end{subfigure} \hfill
    \begin{subfigure}[t]{0.16\textwidth}
		\includegraphics[width=\textwidth]{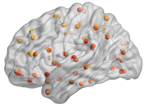}
		\caption{$A_7$}
	\end{subfigure}	\hfill
    \begin{subfigure}[t]{0.16\textwidth}
		\includegraphics[width=\textwidth]{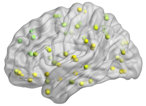}
		\caption{$A_8$} 
	\end{subfigure}\hfill
    \begin{subfigure}[t]{0.16\textwidth}
		\includegraphics[width=\textwidth]{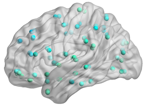}
		\caption{$A_9$}
	\end{subfigure}\hfill
    \begin{subfigure}[t]{0.16\textwidth}
		\includegraphics[width=\textwidth]{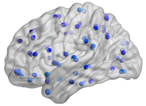}
		\caption{$A_{10}$}
	\end{subfigure}	\hfill
    \begin{subfigure}[t]{0.16\textwidth}
		\includegraphics[width=\textwidth]{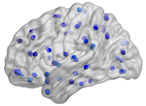}
		\caption{$A_{11}$}
	\end{subfigure}	\hfill  	
	\begin{subfigure}[t]{0.16\textwidth}
		\includegraphics[trim={5cm 1cm 4cm 1cm},clip, width=\textwidth]{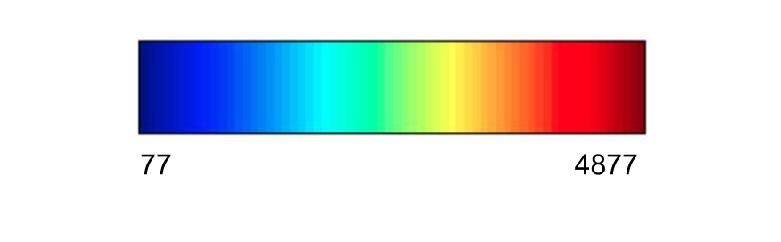}
	\end{subfigure}

    \begin{subfigure}[t]{0.16\textwidth}
		\includegraphics[width=\textwidth]{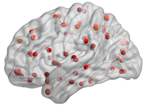}
		\caption{$D_1$}
	\end{subfigure}	\hfill
    \begin{subfigure}[t]{0.16\textwidth}
		\includegraphics[width=\textwidth]{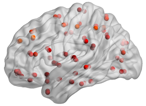}
		\caption{$D_2$} 
	\end{subfigure}\hfill
    \begin{subfigure}[t]{0.16\textwidth}
		\includegraphics[width=\textwidth]{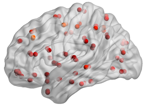}
		\caption{$D_3$}
	\end{subfigure}\hfill
    \begin{subfigure}[t]{0.16\textwidth}
		\includegraphics[width=\textwidth]{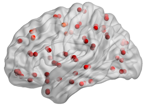}
		\caption{$D_4$}
	\end{subfigure}	\hfill
    \begin{subfigure}[t]{0.16\textwidth}
		\includegraphics[width=\textwidth]{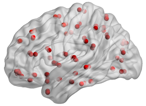}
		\caption{$D_5$}
	\end{subfigure}	\hfill  	
	\begin{subfigure}[t]{0.16\textwidth}
		\includegraphics[width=\textwidth]{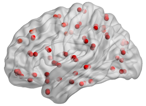}
		\caption{$D_6$}
	\end{subfigure} \hfill
    	\begin{subfigure}[t]{0.16\textwidth}
		\includegraphics[width=\textwidth]{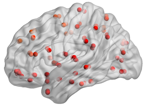}
		\caption{$D_7$}
	\end{subfigure}	\hfill
    \begin{subfigure}[t]{0.16\textwidth}
		\includegraphics[width=\textwidth]{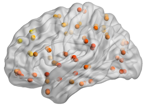}
		\caption{$D_8$} 
	\end{subfigure}\hfill
    \begin{subfigure}[t]{0.16\textwidth}
		\includegraphics[width=\textwidth]{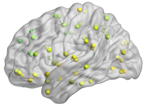}
		\caption{$D_9$}
	\end{subfigure}\hfill
    \begin{subfigure}[t]{0.16\textwidth}
		\includegraphics[width=\textwidth]{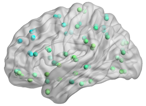}
		\caption{$D_{10}$}
	\end{subfigure}	\hfill
    \begin{subfigure}[t]{0.16\textwidth}
		\includegraphics[width=\textwidth]{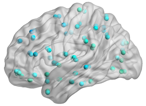}
		\caption{$D_{11}$}
	\end{subfigure}	\hfill  	
	\begin{subfigure}[t]{0.16\textwidth}
		\includegraphics[trim={5cm 1cm 4cm 1cm},clip, width=\textwidth]{colorbar_betcen.jpg}
        \caption{}
	\end{subfigure}
	\caption{Overlay of average node betweenness centrality values on a human brain template.}
    \label{fig:brain_betcen}
\end{figure*}

\begin{figure*}[t]
	\centering
    \includegraphics[trim={4.5cm 0cm 4.2cm 1cm},clip, width=.85\textwidth]{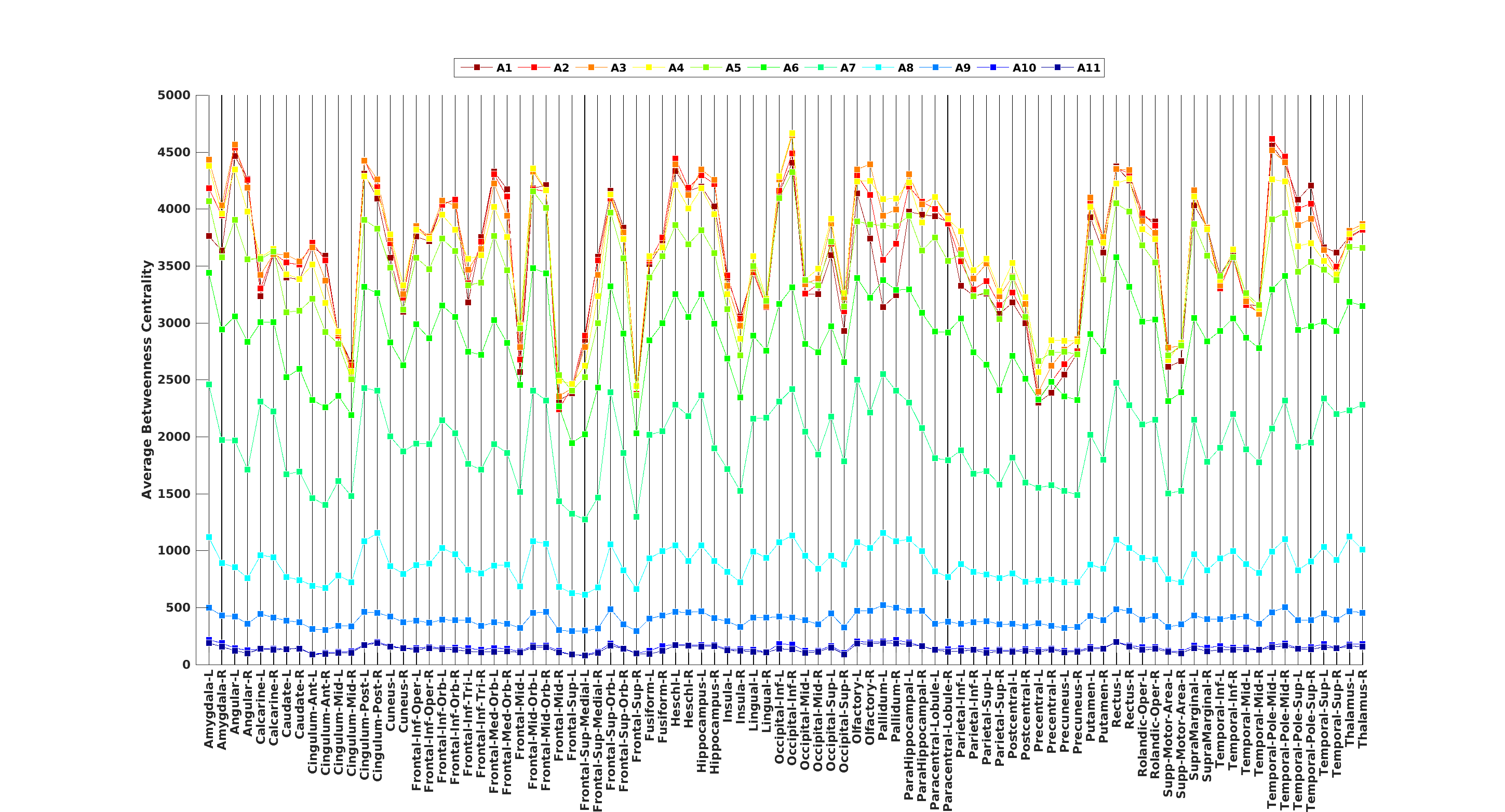}			
	\caption{Average node betweenness centrality obtained for subbands $[A_1, A_{11}]$ for anatomical regions}
    \label{fig:node_bc_point_app}
\end{figure*}

\begin{figure*}[t!]
	\centering
    \includegraphics[trim={4.5cm 0cm 4.2cm 1cm},clip, width=.85\textwidth]{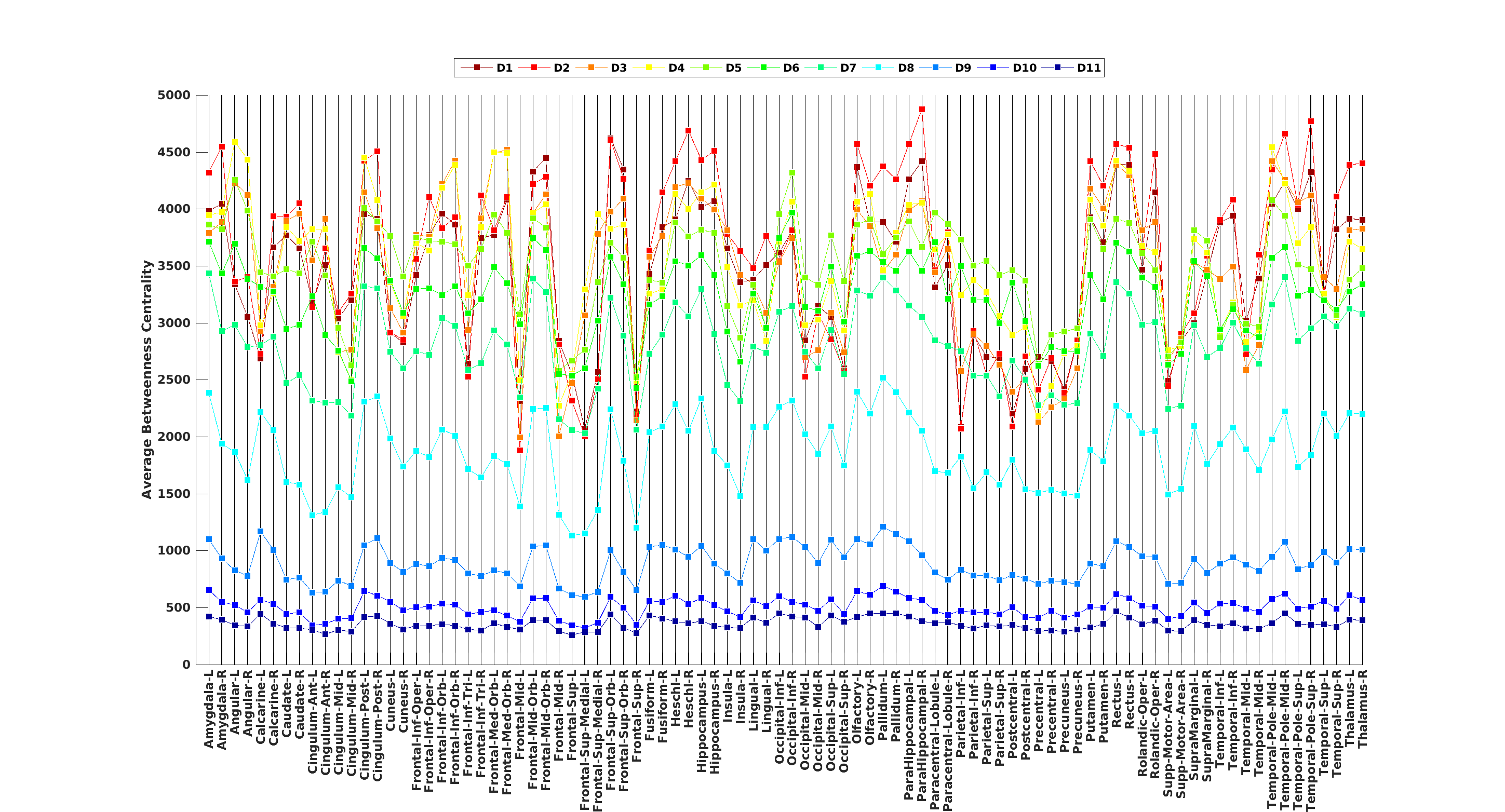}			
	\caption{Average node betweenness centrality obtained for subbands $[D_1, D_{11}]$ for anatomical regions}
    \label{fig:node_bc_point_det}
\end{figure*}

Figure \ref{fig:brain_degree} shows the average node out-degrees on a human brain template for all subbands. In these plots, average node out-degrees are quite low and and similar to each other at high decomposition levels. On the other hand, for subbands $A_1 - A_4$, regions mainly in frontal lobe, especially in supplementary motor area, precentral regions, bilateral superior frontal regions and bilateral medial frontal regions (see Figure \ref{fig:node_degree_point_app}) have large out-degrees. We also observe large out-degrees in precuneus.  On the contrary,  olfactory gyrus, and bilateral pallidum and rectus regions have relatively smaller out-degrees. In lower subbands, we observe approximately same number of node out-degrees for all regions.

Similarly, we observe large out-degrees mainly in frontal lobe for subbands $D_1 - D_4$. Also,  superior frontal gyri, medial frontal gyri precentral and precuneus regions have large out-degrees (see Figure \ref{fig:node_degree_point_det}). Nodes with small out-degrees for these subbands correspond to bilateral olfactory, pallidum, rectus regions together with amygdala and bilateral medial temporal poles. Comparison of Figure \ref{fig:node_degree_point_det} and Figure \ref{fig:node_degree_point_app} reveals that the behavior of node degree distribution, specifically, the location of peaks are quite similar for the approximation and detail parts. This fact indicates that the node degrees of brain regions are represented in both approximation and detail parts, consistently. 

\subsubsection{Connectivity Analysis by Node Strength}

Node strength represents the sum of the arc weights connected to a node where node out-strength is the sum of weights of outward links. In this subsection, we compute the node out-strengths for each region and for each subband. Then, we overlay them on a brain template. Figure \ref{fig:brain_strength} represents node out-strengths for each subband. It can be observed that as the decomposition level increases, the values of node out-strength decreases and approaches to each other within the subband. This behavior is observed in both approximation and detail parts of fMRI signal. 

In order to further analyze the node strength values of anatomical regions, we plot them for approximation (Figure \ref{fig:node_strength_point_app}) and detail (Figure \ref{fig:node_strength_point_det}) parts separately. Regions with maximum node strength in subbands $[A_1, A_5]$ are right medial frontal, inferior and medial occipital (left) and bilateral middle frontal orbital gyri while regions with minimum node-strength are bilateral superior frontal gyri, precuneus (left), middle frontal (left). Although there is a great resemblance at the location of the peaks, relative node out-strength values of regions change within the subbands. For example, the highest node out-strength value in $A_1$ is observed  at right medial frontal, inferior region; whereas the highest peak in $A_4$ is in left parietal inferior region. Note that for subbands $[A_6 , A_{11}]$, all regions have similar out-strength values, yielding almost a flat distribution. This observation may be attributed to the negligible amount of information conveyed in subbands with higher levels than $l=7$.

Figure \ref{fig:node_strength_point_det} shows  node out-strength behavior for detailed parts of mesh networks. In this case, regions with maximum node strength values in subbands $[D_1, D_6]$ are bilateral middle frontal orbital, bilateral superior frontal orbital and bilateral angular gyri. On the other hand, regions with minimum node strength are superior frontal medial (left), superior frontal (right), precentral (left), precuneus (left) and postcentral (left) regions. For subbands $[D_6 , D_{11}]$, all regions have similar and low strength values. Note that, the detail parts distribute the node out-strength across the bands more than the approximation parts, where most of the node out-strength is observed in subbands $D_1-D_{11}$.

We also analyze the total out-strength of mesh networks by summing out-strengths over all nodes. The results obtained for total strength in Figure~\ref{fig:total_strength} show that as the decomposition level increases, total strength of approximation part decreases monotonically. On the other hand, total strength of detail parts shows a slight increase up to level 4, and then decreases with increasing level of decomposition. These results are also compatible with classification accuracy of each subband, given in Table~\ref{tab:single-subband}. 


\begin{figure*}[t]
	\centering
	\begin{subfigure}[t]{0.4\textwidth}
		\includegraphics[width=\textwidth]{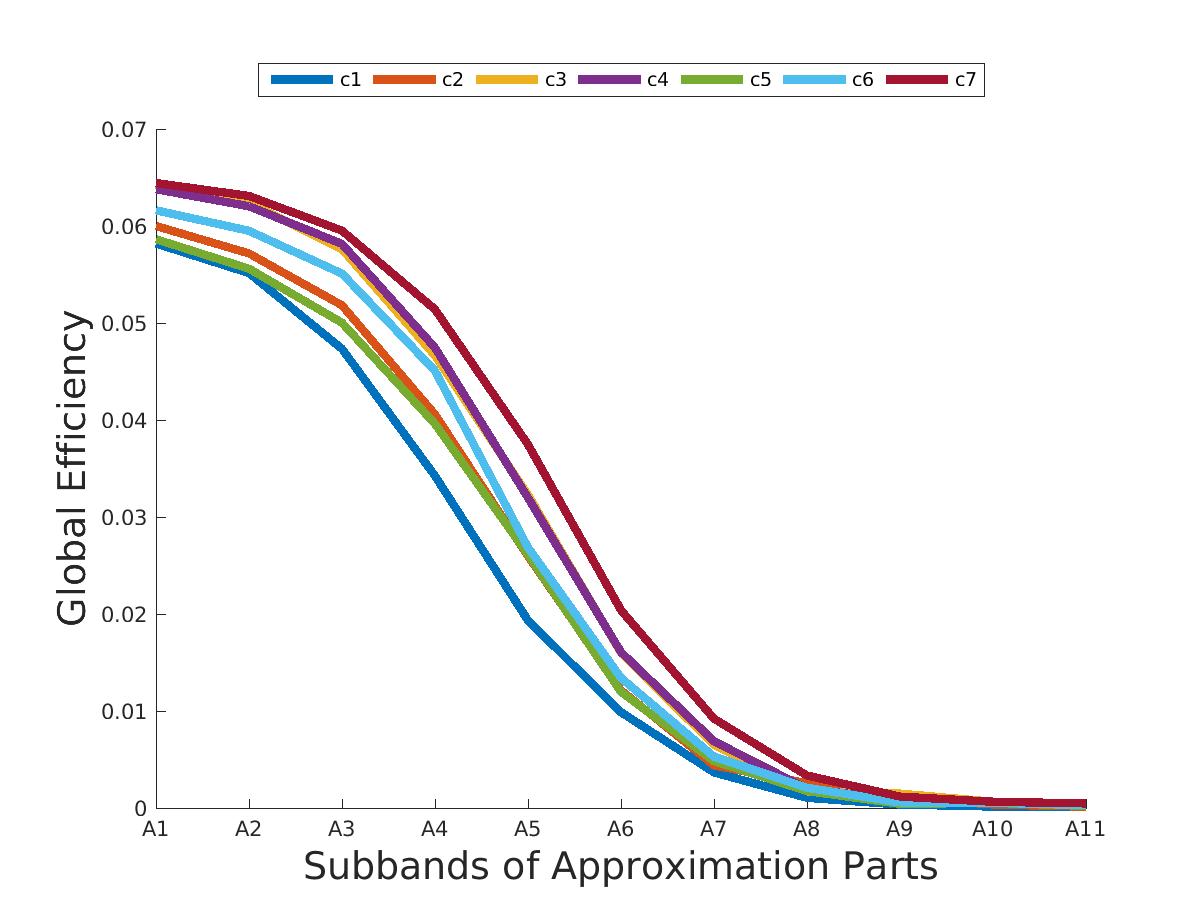}
		\label{fig:globef_app}
	\end{subfigure}%
    \begin{subfigure}[t]{0.4\textwidth}
		\includegraphics[width=\textwidth]{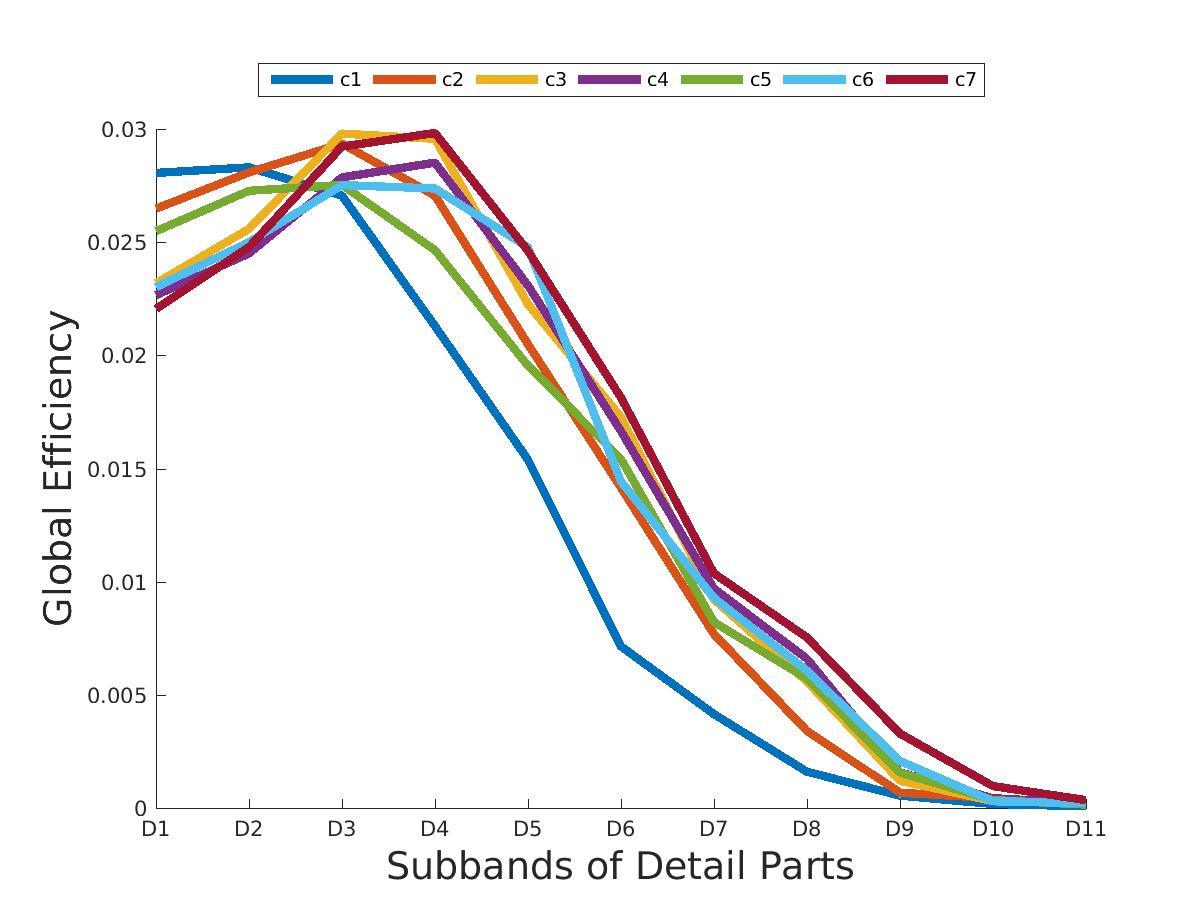}
		\label{fig:globef_det}
	\end{subfigure}
	\caption{Average global efficiency of all mesh networks formed for each task category, over the subbands, for approximation and detail parts.}
    \label{fig:globef}
\end{figure*}

\subsubsection{Connectivity Analysis by Node Betweenness Centrality}

Node betweenness centrality represents how central a node is by considering the fraction of all shortest paths containing that node. In other words, nodes with large betweenness centrality take part in large number of shortest paths. Nodes with large betweenness centrality are candidates to be hubs, and expected to be vital for the flow of information through network \cite{Thompson2015}. We computed betweenness centrality of each node and for each subband, separately (see Figure \ref{fig:brain_betcen}). Results reflect that, as the decomposition level increases, betweenness centrality of nodes decreases, and approaches to zero for majority of the nodes. In other words, we have more hub candidates at lower decomposition levels. Notice that, although betweenness centrality values are large for subbands $A_1 - A_4$, we can still observe that different regions have different centrality values within a subband. In these subbands, angular gyrus and inferior occipital gyrus are the main hub candidates.

In order to analyze hub candidates for each subband, we plot node betweenness centrality values for approximation (Figure \ref{fig:node_bc_point_app}) and detail (Figure \ref{fig:node_bc_point_det} parts separately. For subbands $[A_1, A_5]$ hub candidates are bilateral middle temporal pole, inferior occipital (right), angular (right), Heschl gyri and hippocampus (see Figure \ref{fig:node_bc_point_app}). For subbands $[A_6, A_8]$, we have similar hubs inclusing amygdala (left), bilateral middle frontal orbital, bilateral rectus. For $[A_9 , A_{11}]$ all nodes have similar centrality values.

On the other hand, hub candidates for subbands $[D_1, D_2]$ are bilateral parahippocampal gyri, bilateral frontal medial orbital, while for subbands with larger classification accuracy ($[D_3, D_5]$), hub candidates are middle temporal pole (left), bilateral frontal medial orbital, bilateral angular. For larger levels of decomposition, all regions have similar centrality values.

\subsubsection{Connectivity Analysis by Global Efficiency}

Global efficiency measures how efficiently a network exchanges information \cite{Latora2001}. It is a measure that quantifies small-world property of a network, and it is calculated as the average inverse shortest path length in a network. If nodes are disconnected, then the shortest path between them has an infinite length, leading to zero efficiency \cite{Rubinov2010}. In this study, we analyze how global efficiency differs for each task and for each subband.


We also analyze the global efficiency of all graphs formed for each task in a subband. Note that, we used only the positive arc weights in our graphs to find global efficiency. Our results in Figure \ref{fig:globef} show that, as the decomposition level increases, global efficiency monotonically decreases in the approximation parts of data. We observe a similar behavior for the detail parts. However, global efficiency slightly increases in subbands $[D_3, D_6]$ then start to decrease monotonically.  Recall that, efficiency is related to the small-world property of the networks. Based on our results, we can say that our graphs formed at lower decomposition levels are more likely to exhibit small-world property than the ones obtained at higher levels. 


\begin{figure*}[t!]
	\centering
    \includegraphics[width=\textwidth]{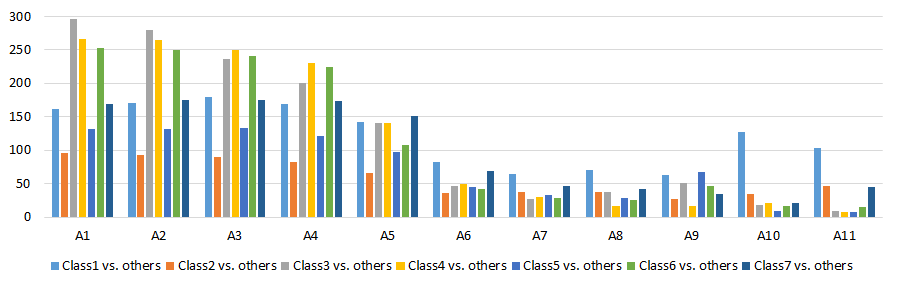}			
	\caption{Maximum $z$ values of class memberships obtained using absolute value two-sample t-test with pooled variance estimate for subbands $[A_1, A_{11}]$. $z$ values are computed under one versus all condition. The bars show the maximum of z-values obtained from the membership values during one-versus-all task classification for each task category.}
    \label{fig:one_vs_all_significance_app}
\end{figure*}

\begin{figure*}[t!]
	\centering
    \includegraphics[width=\textwidth]{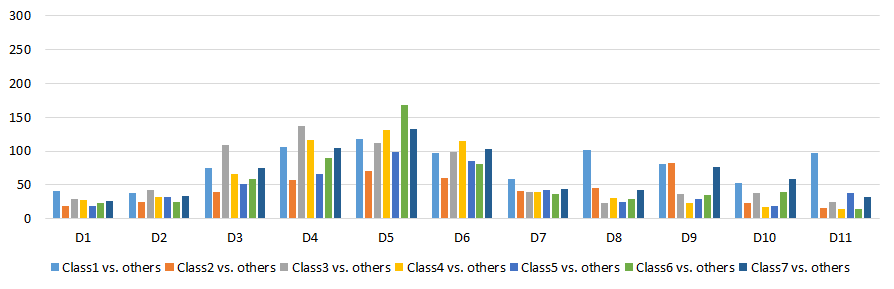}			
	\caption{Maximum $z$ values of class memberships obtained using absolute value two-sample t-test with pooled variance estimate for subbands $[D_1, D_{11}]$. $z$ values are computed under one versus all condition. The bars show the maximum of z-values obtained from the membership values during one-versus-all task classification for each task category.}
    \label{fig:one_vs_all_significance_det}
\end{figure*}

\subsection{Analysis of Class-Discrimination Power of Subbands}


Is there a specific subband or a combination of subbands which represent(s) a specific cognitive task better then the other subbands? Can we represent a cognitive task by combining the mesh networks in selected group of subbands? In this section, we analyze the discriminative properties of base-layer classifiers trained by the mesh networks in different subbands.

Recall that, each base-layer classifier of FSG assigns a class membership vector to a mesh network extracted from a different subband.   We estimate the significance of class membership values using absolute value two-sample t-test with pooled variance estimate. First, we concatenate the class membership vectors obtained at the output of base-layer classifiers under a feature matrix. Then, we divide the feature matrix into two groups, where the first group contains the tasks belonging to a specific task category $c$ and the second group contains the remaining tasks. Then, we compute t-test and obtain $z$ value for each class membership entry as explained in Algorithm \ref{alg:algorithm1}. Notice that, class membership with larger value of $z$ is the one that rejects the null hypothesis that two groups come from the same distribution with more confidence.

Figure \ref{fig:one_vs_all_significance_app} and Figure \ref{fig:one_vs_all_significance_det} denote the significance of class membership values to classify the cognitive tasks in terms of maximum z-values among all the task categories obtained at each subband. High z-values indicate that the corresponding task is well-discriminated in a given subband, whereas the low z-values correspond to classes with inseparable distributions. The general trend of approximation parts in Figure \ref{fig:one_vs_all_significance_app} shows relatively high z-values with a slight decrease, in subbands $[A_1- A_5]$, which means that these subbands are powerful to discriminate all the categories. z-values for task category \textit{Language} (3), \textit{Motor}(4) and \textit{Social} (6) are relatively high (above 200) compared to that of the \textit{Emotion (1), Gambling (2), Relational (5) }and \textit{Working Memory (7)} categories. Behavior of detail parts in Figure \ref{fig:one_vs_all_significance_det} are rather different compared to approximation parts. z-values for task categories \textit{Emotion (1), Motor (4), Relational (5)  } and \textit{Social (6)} reach to maximum z-values at $D_5$ and then decrease gradually, as the level of subbands increases. Task categories \textit{Gambling (2)} and \textit{Language (3)} have the highest z-values at $D_9$ and  $D_4$, respectively. The analysis of the z-values shows that by combining different groups of subbands it is possible to improve the representation power of each task category. For example, task category \textit{Emotion (1)} can be represented in subbands $[A_1- A_5]$, $A_{10}$,  $A_{11}$, $D_{4}$, $D_{5}$, $D_{8}$ and $D_{11}$.  Task category \textit{Gambling (2)} has relatively small z-values in all subbands. However, discriminative power is scattered over the subbands $[A_1- A_5]$, $A_{11}$, $D_{4}$, $D_{5}$, $D_{6}$ and $D_{9}$ for this task. Task category \textit{Language (3)} has the highest z-values among all categories in subbands $[A_1- A_5]$, $D_{3}$ and $D_{4}$. Therefore, these subbands carry significant information to discriminate this category from the rest. Task category \textit{Motor (4)} and \textit{Social (6)} have similar z-values. It is interesting to note that there is a jump of z-value for category 6, in subband $D_5$. Similarly, task categories \textit{Relational (5)} and \textit{Working Memory (7)} have relatively large z-values, in subband $[A_1- A_5]$ and $D_5$. 
In summary, variations of z-values with respect to the subbands, show the discriminative power of each subband for a specific task. This observation indicates that the tasks are best represented in different subband combinations.

\subsection{Analysis of Diversity of Classifiers}

In order to analyze the degree of collaboration and cooperation among the base-layer classifiers trained with the mesh networks of different subbands, we compute popular diversity measures, suggested in Diversity Toolbox \cite{Kuncheva2004}. We analyze the relationship between the state-of-the-art diversity measures and classification accuracy of the proposed HMMN framework.

Recall that, each base-layer classifier outputs a membership value for a mesh network to belong to one of the $c$ task categories. While measuring the diversity of classifiers, we use the oracle outputs which assign  correct/incorrect decision of classifiers. If the membership value is greater than 0.5, then we assume that the input mesh network is generated by this class and assign 1 to the output. Otherwise, the value of the output is 0. The collaboration among the base-layer classifiers are quantified by both pairwise and non-pairwise diversity measures. Pairwise measures include the disagreement measure (\textbf{Disagr.}), Yule's Q statistic (\textbf{Mean Q}), the correlation coefficient (\textbf{Mean} $\boldsymbol{\rho}$); and non-pairwise measures include the entropy measure of an ensemble (\textbf{Ent.}), the measure of inter-rater agreement ($\boldsymbol{\kappa}$), Kolhavi-Wolpert variance (\textbf{KW}) and the measure of difficulty ($\boldsymbol{\theta}$). We average the values of pairwise measures across all pairs of base-layer classifiers. 

\begin{algorithm}[t!]
  \caption{Finding significance of class membership values using  absolute value two-sample t-test with pooled variance estimate}\label{alg:algorithm1}
  \begin{algorithmic}[1]
    \Require Feature matrix of class membership values $\mathcal{M}$ for all tasks ($\forall q$)
    \Require Class label vector $\mathcal{C}$ of all tasks ($\forall q$)
    
    \Ensure Cell array $\mathcal{T}$ containing $z_c$ values, for all classes ($\forall c$) such that $\mathcal{T}\{c\} = z_{c}$
      \Statex
      
      \For{$c = 1 \to 7$}
    	\State $groupinds \gets (\mathcal{C} == c)$
        \State $nongroupinds \gets (\mathcal{C} \sim = c)$
        \State $n_1 \gets$ number of tasks with class label $c$
        \State $n_0 \gets$ number of tasks with class label other than $c$
        \State $m_1 \gets \mathcal{M}_{groupinds}$
        \State $m_0 \gets \mathcal{M}_{nongroupinds}$
        \State $z_c \gets \frac{\mu (m_1) - \mu(m_0)}{\sqrt{ \frac{\sigma(m_1)}{n_1} - \frac{\sigma(m_0)}{n_0} } }$  where $\mu(.)$ denotes mean $\sigma(.)$ denotes standard deviation.
        \State $\mathcal{T}\{c\} \gets z_{c}$
    \EndFor
     
      \end{algorithmic}
\end{algorithm}

Table \ref{tab:diversity-overlap},shows the diversity measures for the classifier groups starting from combining subbands in the first two levels, $A_2$, $D_2$ and $A_1$ and then, adding pairs of approximation and detail parts at each level.  Higher values of \textbf{Disagr.}, \textbf{Ent.}, \textbf{KW} and lower values of $\boldsymbol{\kappa}$, \textbf{Mean Q}, \textbf{Mean} $\boldsymbol{\rho}$, $\boldsymbol{\theta}$ correspond to more diversity of classifiers. In Table \ref{tab:diversity-overlap}, the arrows $(\uparrow)$ or $(\downarrow)$ denote that the diversity is greater if the measure is higher or lower, respectively. Note that the diversity measures consistently show a slight decay in diversity, then it gets better as we add more levels reaching to the best diversity, when we include the classifiers trained in all subbands. This observation shows that all of the subbands carry complementary information about the cognitive tasks.

\begin{table*}[htbp]
\centering
\caption{Diversity measures obtained for ensemble of classifiers trained with mesh arc weights. The arrow specifies whether diversity is larger if the measure is smaller $(\downarrow)$ or larger $(\uparrow)$.}
\label{tab:diversity-overlap}
\renewcommand{\arraystretch}{0.9}
\begin{tabular}{cccccccccc}

& \multicolumn{3}{c}{\textbf{Pairwise}} & & \multicolumn{4}{c}{\textbf{Non-pairwise}} \\
\cmidrule[.8pt]{2-4}
\cmidrule[.8pt]{6-9}
\textbf{Level} & \textbf{Disagr.$(\uparrow)$} & \textbf{Mean Q$(\downarrow)$} & \textbf{Mean $\rho$$(\downarrow)$}& & \textbf{Ent.$(\uparrow)$} & \textbf{K$(\downarrow)$} & \textbf{KW$(\uparrow)$}  & \textbf{$\theta$$(\downarrow)$} \\ \hline
 \textbf{11} & 0.47 & 0.37 & 0.14 & & 0.80 & 0.03  & 0.22  & 0.02 \\
 \textbf{10} & 0.46 & 0.38 & 0.12 & & 0.74 & 0.02  & 0.22  & 0.02 \\
 \textbf{9}  & 0.42 & 0.40 & 0.12 & & 0.64 & 0.02  & 0.20  & 0.02 \\
 \textbf{8}  & 0.36 & 0.45 & 0.14 & & 0.50 & 0.03  & 0.17  & 0.02 \\
 \textbf{7}  & 0.27 & 0.60 & 0.19 & & 0.35 & 0.07  & 0.12  & 0.02 \\
 \textbf{6}  & 0.26 & 0.65 & 0.21 & & 0.34 & 0.07  & 0.12  & 0.02 \\
 \textbf{5}  & 0.17 & 0.73 & 0.25 & & 0.21 & 0.10  & 0.07  & 0.02 \\
 \textbf{4}  & 0.19 & 0.71 & 0.27 & & 0.25 & 0.07  & 0.08  & 0.02 \\
 \textbf{3}  & 0.24 & 0.68 & 0.28 & & 0.33 & 0.04  & 0.10  & 0.03 \\
 \textbf{2}  & 0.32 & 0.58 & 0.27 & & 0.48 & 0.09 & 0.11  & 0.04          \\ \hline

\end{tabular}
\end{table*}

Comparison of Table \ref{tab:wavelet-overlap} and \ref{tab:diversity-overlap} shows the relationship between the diversity measures and classification accuracy. Note that, performances and diversities  increase proportionally, as we add classifiers to the ensemble. These results show that the information about the task categories released in each subband complements each other. Therefore, fusing the base-layer classifier results in boosted performances.

\section{Summary and Conclusion}
We propose a new framework, called Hierarchical Multi-resolution Mesh Networks (HMMNs) for decoding and analysis of cognitive tasks, based on course-to-fine representation of brain connectivity networks, extracted from multi-resolution fMRI data. The proposed approach integrates multi-resolution analysis of fMRI signals, graph embedding of mesh networks, and fusion of fuzzy decisions of classifiers under a unified approach. 

We observe that decomposing the fMRI signal into multiple resolutions also decomposes  task specific information sheltered in the original signal and accentuates the task categories in subsets of subbands.  Moreover, we observe that ensembling a set of local meshes forms a brain network, which represents the locality and distributivity properties of brain under the same framework. The mesh networks extracted in different resolutions enable us to represent the brain  connectivity  in  multiple resolution. This representation is capable of discriminating the cognitive tasks in different resolutions much better compared to pairwise connectivity or average region time series. 

Our analyses on network topology in multiple resolutions show that although low level resolutions carry substantial representation power, high decomposition levels consist of supplementary information about the cognitive tasks. This fact is also observed during the analysis of discrimination power of each subband with respect to cognitive tasks. In other words, discriminative power of subbands changes depending on the cognitive task.  Finally, we observe that diversity of classifiers increases, as we train more and more subbands at the base-layer. As a result, fusion of decisions of base-layer classifiers boosts the performance of the individual classifiers for brain decoding. 

As a future work, a better approach for HMMNs may be to partition the fMRI data into homogeneous regions and replace the anatomic regions by the homogeneous regions to form the representative time series. Mesh networks formed among the homogeneous regions are expected to have a better representation power compared to the average time series of anatomic regions. A further research direction is to compare resting state fMRI data between patients and control groups using our proposed framework. 

\section*{Acknowledgments}
This work was supported by CREST, JST, and TUBITAK Project No 116E091. Itir Onal Ertugrul was supported by TUBITAK.


\bibliography{mybibfile}

\end{document}